\documentclass{l4dc2024}

\usepackage{cite}
\usepackage{amsfonts}
\usepackage{algorithmic}
\usepackage{enumitem}
\usepackage{pifont}
\usepackage{times}
\newcommand{\diff}{\mathrm{d}}
\usepackage{diagbox}
\newcommand{\abovegap}{1ex}
\newcommand{\belowgap}{1ex}

\newtheorem{assumption}{Assumption}

\title{Learning ``Look-Ahead" Nonlocal Traffic Dynamics in a Ring Road} 

\author{%
 \Name{Chenguang Zhao} \Email{czhao704@connect.hkust-gz.edu.cn}\\
 \addr Thrust of Intelligent Transportation, The Hong Kong University of Science and Technology (Guangzhou)
 \AND
 \Name{Huan Yu}\thanks{corresponding author}\Email{ huanyu@ust.hk} \\
 \addr Thrust of Intelligent Transportation, The Hong Kong University of Science and Technology (Guangzhou)\\
 Department of Civil and Environmental Engineering, The Hong Kong University of Science and Technology%
}
\begin{document}

\maketitle
\vspace{-1em}
\begin{abstract}%
The macroscopic traffic flow model is widely used for traffic control and management. To incorporate drivers' anticipative behaviors and to remove impractical speed discontinuity inherent in the classic Lighthill–Whitham–Richards (LWR) traffic model, nonlocal partial differential equation (PDE) models with ``look-ahead" dynamics have been proposed, which assume that the speed is a function of weighted downstream traffic density. However, it lacks data validation on two important questions: whether there exist nonlocal dynamics, and how the length and weight of the ``look-ahead" window affect the spatial temporal propagation of traffic densities. In this paper, we adopt traffic trajectory data from a ring-road experiment and design a physics-informed neural network to learn the fundamental diagram and look-ahead kernel that best fit the data, and reinvent a data-enhanced nonlocal LWR model via minimizing the loss function combining the data discrepancy and the nonlocal model discrepancy. Results show that the learned nonlocal LWR yields a more accurate prediction of traffic wave propagation in three different scenarios: stop-and-go oscillations, congested, and free traffic. We first demonstrate the existence of ``look-ahead" effect with real traffic data.  The optimal nonlocal kernel is found out to take a length of around 35 to 50 meters, and the kernel weight within 5 meters accounts for the majority of the nonlocal effect. Our results also underscore the importance of choosing a priori physics in machine learning models. 
\end{abstract}


\section{Introduction}

Macroscopic traffic flow models describe the dynamics of aggregated traffic states, density, speed, and flow, by partial differential equations (PDEs). It serves as the basis for various traffic management tools, such as those alleviating congestion~\citep{yu2022traffic}, improving throughput~\citep{smith2019traffic}, and reducing emissions~\citep{rodriguez2021coupled}. The first and widely used traffic flow model is the Lighthill–Whitham–Richards (LWR) model~\citep{lighthill1955kinematic,richards1956shock}, which describes the dynamics of density as a first-order hyperbolic PDE. It is derived from the conservation law and assumes that speed is decided by a static function of local density, also known as the fundamental diagram. Despite the simplicity, it causes shock waves in finite time with smooth initial conditions. The discontinuity of speed is not realistic in real traffic as the acceleration becomes infinite at the shock wave. To avoid discontinuity, the nonlocal LWR model has been proposed in which the speed is decided by a weighted average of downstream  or upstream traffic density within a finite length ~\citep{blandin2016well}. Human drivers have ``look-ahead'' of downstream traffic because they can anticipate and react not only to local traffic as in preceding leader vehicles, but may also to traffic conditions further downstream within the field of vision. The nonlocal effect can be amplified for traffic on a ring road, as shown in Fig.~\ref{fig:introduction}, where the span of drivers' vision is largely enhanced due to the curvature of the ring road, compared to straight roads. When there are connected automated vehicles that are controlled based on upstream vehicle information received from vehicle-to-vehicle communication, the traffic will present ``look-behind'' nonlocal property.

Theoretical analysis has proved that both ``look-ahead'' and ``look-behind'' nonlocal extensions of LWR generate smooth solutions under certain assumptions on the initial condition, boundary condition, and model parameters~\citep{karafyllis2022analysis}. Besides, ``look-behind'' nonlocal controllers will increase traffic capacity~\citep{karafyllis2022analysis}. Some analytical properties of the nonlocal LWR model have also been discussed, such as well-posedness~\citep{goatin2016well}, controllability and stability~\citep{bayen2021boundary,huang2022stability}, the vanishing nonlocality limit, \citep{colombo2019singular,keimer2019approximation}. However, to the best of our knowledge, there has been little research characterizing nonlocal traffic dynamics with field data. It remains an open question whether nonlocal dynamics exist in real traffic, and if so, what form the nonlocal function and the fundamental diagram can be validated with data. In this paper, we carefully choose the trajectory data of human drivers collected from a ring-road setting and present the first result on analyzing the nonlocal traffic dynamics, particularly ``look-ahead" from the data.  

Real traffic data has been collected by many researchers, such as Lagrangian data collected by vehicle sensors~\citep{sun2020scalability,zheng2021experimental} and Euclidean data collected from loop detectors or surveillance cameras~\citep{NGSIM,stern2018dissipation,gloudemans2023I24,krajewski2018highd}.  The data has been used to calibrate local LWR models via various methods, such as least square~\citep{dervisoglu2009automatic}, Nelder-Mead method~\citep{kontorinaki2017first,nelder1965simplex}, and genetic algorithm~\citep{mohammadian2021performance}. 
This paper conducts the calibration of nonlocal LWR models with ring-road traffic data. It is a more challenging task than the local ones. In local LWR models, the fundamental diagram reflects the dependence of speed and local density, which are directly measurable from detectors. While in nonlocal LWR models, the nonlocality results in coupling between nonlocal kernel function and local density-speed relation. The coupling is then embedded into a dynamic spatial-temporal propagation of traffic density, which makes it very difficult to identify from the data.  Recently developed Physics-informed deep learning (PIDL)~\citep{raissi2019physics}, a physics-constrained learning approach, has been adopted to reinvent the local LWR model with data enhancement~\citep{shi2021physics,zhao2023observer}. The PIDL traffic model achieves more accurate prediction than the pure physical model, since a trade-off between data and model can be achieved with a physics-uninformed neural network (NN) to learn the mapping from spatial-temporal location to system state, and a physics-informed NN to learn the system dynamics.  In this paper, we tackle the nonlocality with PIDL and design an NN with physics constraints to learn the optimal fundamental diagram and kernel function.

The main contribution of this paper lies in first characterizing, learning, and analyzing the ``look-ahead" traffic dynamics that best fit  experiment traffic data. We adopt PIDL to learn the optimal fundamental diagram and kernel function. The NN is optimized to minimize a loss function of three components: a data loss that reflects the discrepancy between learned density and ground truth measurement, a physics dynamics loss that evaluates that discrepancy between learned dynamics and the nonlocal LWR  model, and a physics static loss designed to satisfy  constraints on the fundamental diagram the kernel function for  well-posedness of the nonlocal LWR model.
Based on the learned kernel function and fundamental diagram, we find that the nonlocal LWR yields a more accurate estimation of traffic dynamics, i.e., the propagation of traffic waves.

\section{Ring-road data and nonlocal traffic flow model}

\subsection{Ring-road traffic data}

\begin{figure}[t]
    \centering    \includegraphics[width=0.75\linewidth]{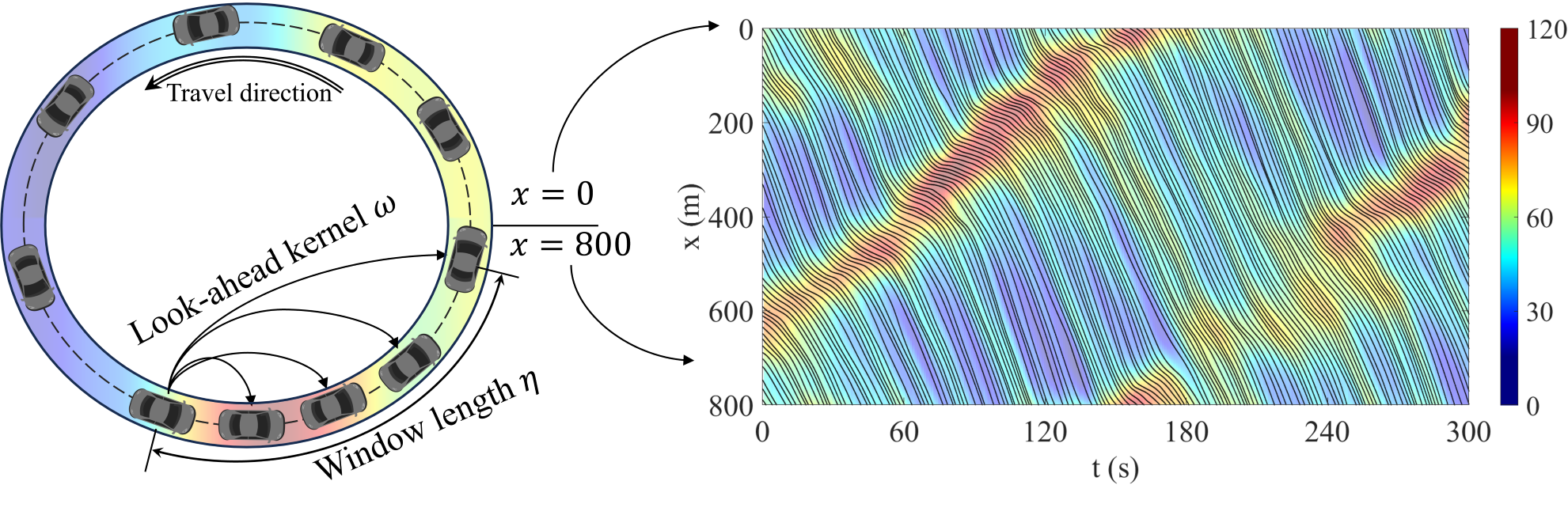}
    \vspace{-2em}
    \caption{Ring road traffic dynamics. Left: ring road experiment setting. Right: vehicle trajectory (black line) and spatial-temporal propagation of density.}
    \label{fig:introduction}
    \vspace{-1.5em}
\end{figure}

Several experiments have been conducted to collect traffic data on ring-roads. In~\citep{sugiyama2008traffic}, the authors arrange 22 vehicles on a ring road with a circumference of 230 m. In~\citep{stern2018dissipation}, experiments are conducted with 22 vehicles on a ring road with a circumference of 230 m. The most up-to-date experiment  \citep{zheng2021experimental} collects data from 40 vehicles traveling on an 800-meter  circumference ring road, which we will use in this paper. 
Each vehicle is equipped with high precision  GPS to record its location and speed with a frequency of 10 Hz. Measurement error via the GPS is of $\pm 1$ m for location and $\pm 1$ km/h for density. To reconstruct the macroscopic density and speed from recorded vehicle trajectories, we  adopt kernel density estimation (KDE) with a Gaussian kernel~\citep{fan2013data,parzen1962estimation}. The reconstructed states are in a discrete domain $\mathcal{G} = \left\{(t_i,x_j)|i=0,\cdots,N_t-1; x=0,\cdots,N_x-1\right\}$.  We take the cell interval being  $\Delta t = 1$ s  and $\Delta x = 1$ m. The reconstructed traffic density is visualized in Fig.~\ref{fig:introduction}.

\subsection{Nonlocal traffic flow model}
We consider the macroscopic traffic dynamics on a ring road of length $L>0$. Based on the conservation law of vehicle numbers, we have the following hyperbolic PDE: 
\begin{equation}
    \setlength{\abovedisplayskip}{\abovegap}
    \setlength{\belowdisplayskip}{\belowgap}
    \partial_t \rho(x,t) + \partial_x (\rho(x,t)v(x,t)) = 0,
\end{equation}
where $\rho(x,t)$ and $v(x,t)$ represent the density and speed at time $t\ge 0$ and location $0\le x \le L$ respectively. On a ring road, we have periodic boundary conditions, i.e.,  $\rho(x+L,t) = \rho(x,t) $. The LWR model assumes that speed $v(x,t)$ is dependent on the local density $\rho(x,t)$ by the fundamental diagram $V(\rho):\mathbb{R}\to \mathbb{R}$, i.e., $v(x,t) = V(\rho(x,t))$. The evolution of traffic density is then given as:
\begin{equation}
    \setlength{\abovedisplayskip}{\abovegap}
    \setlength{\belowdisplayskip}{\belowgap}
    \partial_t \rho(x,t) + \partial_x \left( \rho (x,t)V(\rho(x,t)) \right)= 0.
\end{equation}
Considering the physics constraints, we have Assumption~\ref{assumption:V} on the fundamental diagram $V(\rho)$.
\begin{assumption}\label{assumption:V}
    The fundamental diagram $V(\rho)$ is a non-negative, non-increasing function.
\end{assumption}

To describe the dependence between speed and density, some closed-form functions with parameters have been adopted as the fundamental diagram.   \citep{greenshields1935study} takes the first step to calibrate the fundamental diagram as a linear function:
\begin{equation}
   \setlength{\abovedisplayskip}{\abovegap}
    \setlength{\belowdisplayskip}{\belowgap}
    V(\rho) = v_f  \left(1-{\rho}/{\rho_m}\right), \label{eq:FD greenshields}
\end{equation}
with $v_f>0$ and $\rho_m>0$ being parameters representing free speed and maximum density.  In~\citep{underwood1961speed}, the fundamental  diagram  is:
\begin{equation}\label{eq:FD exp}
    \setlength{\abovedisplayskip}{\abovegap}
    \setlength{\belowdisplayskip}{\belowgap}
    V(\rho) = v_f \exp \left(-{\rho}/{\rho_c}\right), 
\end{equation}
with the two parameters $v_f>0$ and $\rho_c>0$ being free speed and critical density respectively. \citep{drake1966statistical} describes the fundamental diagram via:
\begin{equation}\label{eq:FD exp2}
\setlength{\abovedisplayskip}{\abovegap}
    \setlength{\belowdisplayskip}{\belowgap}
    V(\rho) = v_f \exp \left(-(\rho/\rho_c)^2/2\right). 
\end{equation}
Besides using closed-form functions with parameters that have explicit physical meaning, researchers have also focused on getting the fundamental diagram via data-driven methods, such as generalized least square method in~\citep{qu2015fundamental}, machine learning in~\citep{shi2021physics}, and Gaussian process in~\citep{cheng2022bayesian}. However, all these methods focus on the calibration of speed and local density. It is unclear how these models fit in with nonlocal dynamics.

Unlike the local LWR model, the ``look-ahead'' nonlocal model assumes that the speed at $(x,t)$ depends not on the local density $\rho(x,t)$, but instead the weighted average of downstream traffic density within a length of $\eta>0$. The speed becomes $v(x,t) = V_{\eta}(\rho_{\eta}(x,t))$, where the nonlocal traffic density takes the form of
\begin{equation}\label{eq:nonlocal rho}
    \setlength{\abovedisplayskip}{\abovegap}
    \setlength{\belowdisplayskip}{\belowgap}
    \rho_{\eta}(x,t) = \int_0^{\eta} \rho(x+y,t)\omega(y) \diff y,
\end{equation}
with $\omega:[0,\eta]\to \mathbb{R}$ being the integral nonlocal kernel function. Similarly, for ``look-behind'' nonlocal LWR, the nonlocal traffic density is a weighted average of upstream traffic. The nonlocal PDE traffic flow model is
\begin{equation}\label{eq:LWR nonlocal}
    \setlength{\abovedisplayskip}{\abovegap}
    \setlength{\belowdisplayskip}{\belowgap}
    \partial_t \rho(x,t) + \partial_x \left(\rho(x,t) V_{\eta}(\rho_{\eta}(x,t))\right)= 0.
\end{equation}
For well-posedness of the nonlocal PDE~\eqref{eq:LWR nonlocal}, besides the same constraints on the fundamental diagram $V_\eta$ as in Assumption~\ref{assumption:V},   we also have constraints on the kernel function $\omega(x)$ as follows.
\begin{assumption}\label{assumption:omega}
    The function $\omega(x)$ is a non-negative, non-increasing function with $\int_0^{\eta} \omega(x) \diff x= 1$.
\end{assumption}

\begin{theorem}\citep{karafyllis2022analysis}\label{theorem:well posedness}
Under Assumption~\ref{assumption:V} on $V_{\eta}$ and Assumption~\ref{assumption:omega} on $\omega$, for every initial condition $\rho (t,0)\in W^{2,\infty}(\mathbb{R})\cap \mathrm{Per}(\mathbb{R})$, 
the initial value problem for the nonlocal LWR~\eqref{eq:LWR nonlocal} has a unique solution 
$\rho(x,t) \in C^1(\mathbb{R}^+\times\mathbb{R})$ with 
$\rho(t,\cdot) \in W^{2,\infty}(\mathbb{R}) \cap  \mathrm{Per}(\mathbb{R}) $ for all $t\ge 0$, 
where $W^{2,\infty}(\mathbb{R})$ is the Sobolev space of $C^1$ functions on $\mathbb{R}$ with Lipschitz derivative, and  $\mathrm{Per}(\mathbb{R})$ is the set of continuous, positive mappings $\rho: \mathbb{R}\to(0,\infty)$ with a period of $L$. 
\end{theorem}
Two commonly used kernel functions take the constant or linear decreasing function form as:
\begin{align}
    &\omega(x) = 1/\eta,\label{eq:omega constant} \\
    &\omega(x) = 2(\eta-x)/\eta^2. \label{eq:omega decrease}
\end{align}
Although these two kernel functions satisfy the conditions in Assumption~\ref{assumption:omega}, it is unclear how well they fit real traffic flow. In this paper, we learn the  kernel function that best fits real traffic data.

\section{Learning fundamental diagram, look-ahead kernel, \& spatial-temporal dynamics}

\begin{figure}[t]
    \centering
    \includegraphics[width=0.65\linewidth]{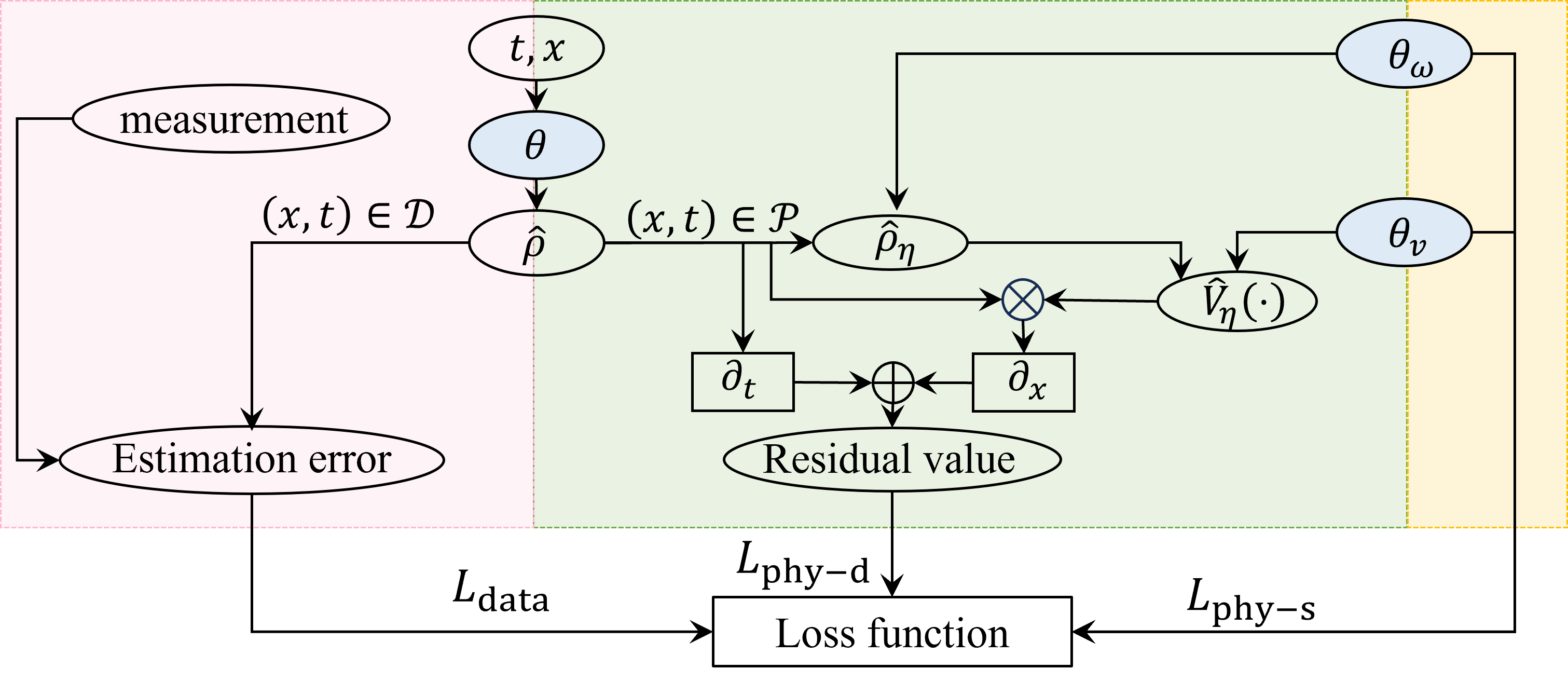}
    \vspace{-1.5em}
    \caption{Diagram of the proposed NN. We use $\mathcal{D}$ to evaluate the data discrepancy between learned density and ground truth, and $\mathcal{P}$ to evaluate the model discrepancy between the learned dynamics and the nonlocal LWR.}
    \label{fig:framework}
      \vspace{-1.5em}
\end{figure}

We give in Fig.~\ref{fig:framework} the constructed NN. There are three parameters to be trained: $\theta$ to learn the spatial-temporal dynamics of density $\rho(x,t)$; $\theta_\omega$ to learn the nonlocal look-ahead kernel $\omega$, and $\theta_v$ to learn the fundamental diagram $V_{\eta}$. We design three loss functions: data loss $L_{\mathrm{data}}(\theta)$ in the red box to describe the discrepancy between the learned density and measurement, physics dynamics loss $L_{\mathrm{phy-d}}$ in the green box to describe the discrepancy between the learned dynamics and nonlocal traffic flow model, and physics static loss $L_{\mathrm{phy-s}}$  in the yellow box to constrain the learned kernel and fundamental diagram to ensure well-posedeness of the nonlocal PDE.
The overall loss function is:
\begin{equation}
   \setlength{\abovedisplayskip}{\abovegap}
    \setlength{\belowdisplayskip}{\belowgap}
    L(\theta,\theta_v,\theta_\omega)= L_{\mathrm{data}}(\theta,\theta_v,\theta_\omega) + L_{\mathrm{phy-d}}(\theta,\theta_v,\theta_\omega) + L_{\mathrm{phy-s}}(\theta_v,\theta_\omega).
\end{equation}
We specify the loss functions as follows.

\subsection{Data loss}

We use an NN with parameter $\theta$ to learn the density at location $x$ and  time $t$ as $\hat \rho(x,t;\theta)$. 
We assume that the initial condition $\rho(0,x)$ is given, and there are $N_l$ loop detectors located at  $x=l_i$ to measure the density $\rho(t,l_i)$. So we have measured traffic density at $\mathcal{D} = \{(0,x_i)|i=0,\cdots N_x-1\} \cup \{(t_i,l_j)|i=0,\cdots,N_t-1;j=0,\cdots N_l-1\}$.
To minimize the difference between learned density $\rho(x,t;\theta)$ and the measurement density $\rho(x,t)$, the  data loss is designed as
\begin{equation}
    \setlength{\abovedisplayskip}{\abovegap}
    \setlength{\belowdisplayskip}{\belowgap}
    L_{\mathrm{data}}(\theta) = \alpha_{-1} L_{\mathrm{initial}}(\theta) + \sum_{i=0}^{N_l-1} \alpha_i L_{\mathrm{detector}} (l_i;\theta),
\end{equation}
where $\alpha_{-1}>0$ and $\alpha_{i}>0$ are coefficients, $L_{\mathrm{initial}}(\theta)$ and $L_{\mathrm{detector}}(\theta)$ are the estimation error on the initial data and loop detector data respectively:
\begin{align}
    L_{\mathrm{initial}}(\theta) &= \frac{1}{N_x} \sum_{i=0}^{N_x}\left(\rho(i\Delta x,0) 
 - \hat \rho(i\Delta x,0;\theta)\right)^2.\\
    L_{\mathrm{detector}}(l_i;\theta) &= \frac{1}{N_t} \sum_{k=0}^{N_t}\left( \rho(t_k,l_i) 
 - \hat \rho(t_k,l_i;\theta)\right)^2.
\end{align}

\subsection{Physics loss}  

\noindent  \textbf{Physics dynamics loss}:    Since NN cannot directly calculate integral in the nonlocal density,  we adopt the numerical scheme with convergence proved in~\citep{karafyllis2022analysis}. A numerically approximated nonlocal density is:
\begin{equation}
\setlength{\abovedisplayskip}{\abovegap}
    \setlength{\belowdisplayskip}{\belowgap}
    \bar{\rho}_{\eta} (t_i,x_j;\theta,\theta_{\omega}) = \sum_{k=0}^{N_{\eta}-1} \rho(t_i,x_j+k\Delta x) \bar{\omega}_k,
\end{equation}
where $N_{\eta} = \eta/\Delta x$, and $\bar{\omega}_k = \int_{k\Delta x}^{(k+1)\Delta x} \omega(s) ds $.   We use a trainable vector $\theta_{\omega}\in \mathbb{R}^{N_{\eta}}$ to learn the kernel $\bar{\omega}$, and the learned kernel $\hat{\bar{\omega}}(\theta_\omega)$ is:
\begin{equation}\label{eq:omega discrete learned}
    \setlength{\abovedisplayskip}{\abovegap}
    \setlength{\belowdisplayskip}{\belowgap}
    \hat{\bar{\omega}}_i(\theta_\omega) = \frac{\theta_{\omega,i}}{\mathbf{1} \cdot \theta_{\omega}}.
\end{equation}
The learned nonlocal density is
\begin{equation}\label{eq:nonlocal rho discrete}
    \setlength{\abovedisplayskip}{\abovegap}
    \setlength{\belowdisplayskip}{\belowgap}
    \hat{\rho}_{\eta} (t_i,x_j;\theta,\theta_{\omega}) = \sum_{k=0}^{N_{\eta}-1} \hat \rho(t_i,x_j+k\Delta x;\theta) \hat{\bar{\omega}}_k(\theta_{\omega}).
\end{equation}
We use a neural network with parameter $\theta_v$ to learn the fundamental diagram $V_{\eta}(\cdot)$. The NN takes density as input and outputs the estimated speed $\hat V_{\eta}(\cdot;\theta_v)$. To evaluate the discrepancy between the learned dynamics and the nonlocal LWR model, we define a residual value as
\begin{equation}
  \setlength{\abovedisplayskip}{\abovegap}
    \setlength{\belowdisplayskip}{\belowgap}
    \begin{aligned}
    f(t_i,x_j;\theta,\theta_v,\theta_{\omega}) = &\partial_t \hat{\rho}(t_i,x_j;\theta) + \partial_x \left( \hat{\rho} (t_i,x_j;\theta) \hat{V}_{\eta}(\hat{\rho}_{\eta}(t_i,x_j;\theta,\theta_{\omega});\theta_v) \right)\\
    = & \partial_t \hat{\rho}(t_i,x_j;\theta) + \partial_x \hat{\rho} (t_i,x_j;\theta) \hat{V}_{\eta}(\hat{\rho}_{\eta}(t_i,x_j;\theta,\theta_{\omega});\theta_v)   \\
    &+  \hat{\rho} (t_i,x_j;\theta)\partial_{\rho}\hat{V}_{\eta}(\hat{\rho}_{\eta}(t_i,x_j;\theta,\theta_{\omega});\theta_v) \partial_x \hat{\rho}_{\eta} (t_i,x_j;\theta,\theta_{\omega}), 
\end{aligned}
\end{equation}
where $ \partial_x \hat{\rho}_{\eta} (t_i,x_j;\theta,\theta_{\omega})$ is given by the same numerical discretization in~\eqref{eq:nonlocal rho discrete} as
\begin{equation}
    \setlength{\abovedisplayskip}{\abovegap}
    \setlength{\belowdisplayskip}{\belowgap}
     \partial_x \hat{\rho}_{\eta} (t_i,x_j;\theta,\theta_{\omega}) = \sum_{k=0}^{N_{\eta}-1}\partial_x \hat{\rho}(x,t+k\Delta x;\theta) \hat{\bar{\omega}}_k(\theta_\omega); 
\end{equation}  
the partial differentiation $\partial_t \hat{\rho} (t_i,x_j;\theta)$, $\partial_x \hat{\rho} (t_i,x_j;\theta)$, and $\partial_{\rho}\hat{V}_{\eta}(\hat{\rho}_{\eta}(t_i,x_j;\theta,\theta_{\omega});\theta_v)$ are calculated via automatic differentiation in Tensorflow. We design a physics loss to minimize the residual value:
\begin{equation}\label{eq:loss physical}
    \setlength{\abovedisplayskip}{\abovegap}
    \setlength{\belowdisplayskip}{\belowgap}
    L_{\mathrm{phy}}(\theta,\theta_v,\theta_{\omega}) = \frac{1}{N_p}\sum_{(i,j)\in P} \left( f(t_i,x_j;\theta,\theta_v,\theta_{\omega}) \right)^2,
\end{equation}
where $\mathcal{P}\in \mathcal{G}$ are temporal-spatial points selected from the whole domain to evaluate the learned dynamics.

\noindent  \textbf{Physics static loss}:  Given the conditions on $\omega(x)$ in Assumption~\ref{assumption:omega}, we have three conditions on $\bar{\omega}_k$: 
\begin{align}
    &\bar{\omega}_k \ge 0, \forall k=0,1,\cdots,N_{\eta}-1, \label{eq:omega bar constraint 1 positive}\\
    &\bar{\omega}_{k+1}\ge \bar{\omega}_k, \forall k=0,1,\cdots,N_{\eta}-2, \label{eq:omega bar constraint 2 decrease}\\
    &\sum_{k=0}^{N_\eta-1} \bar{\omega}_k = 1. \label{eq:omega bar constraint 3 sum1}
\end{align}
By the definition of the learned kernel $\hat{\bar{\omega}}$~\eqref{eq:omega discrete learned}, the constraint~\eqref{eq:omega bar constraint 3 sum1} is automatically satisfied. To meet the constraint~\eqref{eq:omega bar constraint 1 positive}, we define a penalty as:
\begin{equation}
  \setlength{\abovedisplayskip}{\abovegap}
    \setlength{\belowdisplayskip}{\belowgap}
    L_{p,\omega,1} (\theta_\omega)= \sum_{i=0}^{N_{\eta}-1} \left(\min \{\hat{\bar{\omega}}_i(\theta_\omega),0\}\right)^2. 
\end{equation}
To meet the constraint~\eqref{eq:omega bar constraint 2 decrease}, we define a penalty as:
\begin{equation}
\setlength{\abovedisplayskip}{\abovegap}
    \setlength{\belowdisplayskip}{\belowgap}
    L_{p,\omega,2}(\theta_\omega) = \sum_{i=0}^{N_{\eta}-2} \left(\max \{\hat{\bar{\omega}}_{i+1}(\theta_\omega)-\hat{\bar{\omega}}_i(\theta_\omega),0\}\right)^2.
\end{equation}
For the fundamental diagram, given the constraints in Assumption~\ref{assumption:V}, to satisfy $\hat V_{\eta}(\rho;\theta_v)\ge 0$, we design a penalty as
\begin{equation}
    \setlength{\abovedisplayskip}{\abovegap}
    \setlength{\belowdisplayskip}{\belowgap}
    L_{p,v,1} = \sum_{i=0}^{N_{\rho}-1}   \left(\min\left\{ \hat{V}_{\eta} (i\Delta\rho;\theta_v),0\right\}\right)^2 , 
\end{equation}
where $N_{\rho} = \rho_m/\Delta \rho$ with $\rho_m$ being the maximum density.  To make $\hat V_{\eta}(\rho;\theta_v)$ be non-increasing with respect to the density $\rho$, we design a penalty as
\begin{equation}
\setlength{\abovedisplayskip}{\abovegap}
    \setlength{\belowdisplayskip}{\belowgap}
    L_{p,v,2} = \sum_{i=0}^{N_{\rho}-1}   \left( \max\left\{
    \partial_{\rho} \hat V_{\eta}(i \Delta \rho;\theta_v),0
    \right\} \right)^2,
\end{equation}
where $\partial_{\rho} \hat V_{\eta}$ is calculated using automatic differentiation provided in Tensorflow. The physics static loss is a weighted sum of the four penalties:
\begin{equation}
\setlength{\abovedisplayskip}{\abovegap}
    \setlength{\belowdisplayskip}{\belowgap}
    L_{\mathrm{phy-s}} = p_{\omega,1} L_{p,\omega,1} (\theta_\omega)+ p_{\omega,1} L_{p,\omega,2}(\theta_\omega) + p_{v,1} L_{p,v,1}(\theta_v) + p_{v,2} L_{p,v,2}(\theta_v),  
\end{equation}
where $p_{v,1}>0$, $p_{v,2}>0$, $p_{v,3}>0$, $p_{\omega,1}>0$, and $p_{\omega,2}>0$ are coefficients.

\begin{figure}
    \centering
    \begin{minipage}{0.31\linewidth}
    \centerline{\includegraphics[width=1\linewidth]{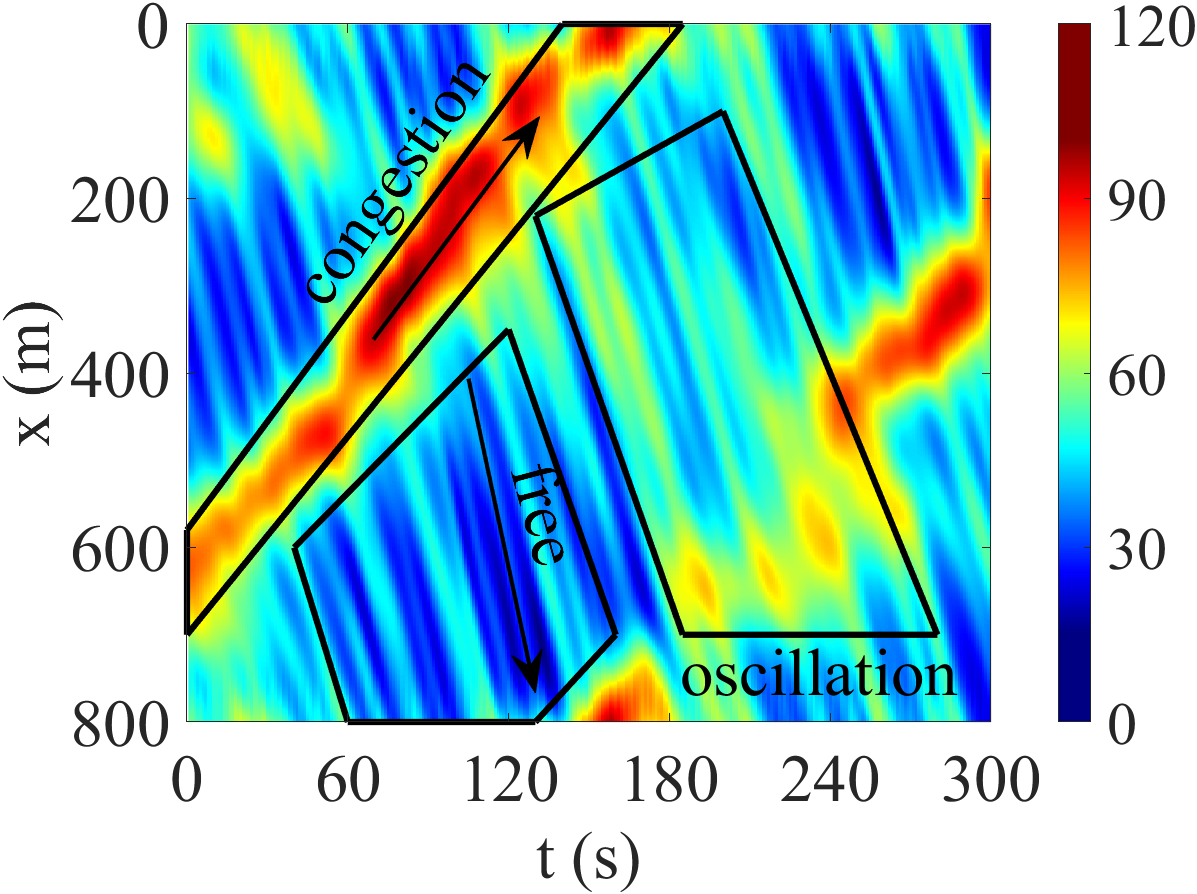}}
    \end{minipage}
    \begin{minipage}{0.68\linewidth}
    \vspace{-1ex}
    There are three significant traffic features:  \\
    A: upstream propagation of congested waves (dark red) \\ B: downstream propagation of free waves (dark blue) \\
    C: stop-and-go oscillation (light red, light blue, and yellow) 
    \end{minipage}
    \\
    Greenshields~\eqref{eq:FD greenshields}\\
    \subfigure[Local]{\includegraphics[width=0.245\linewidth]{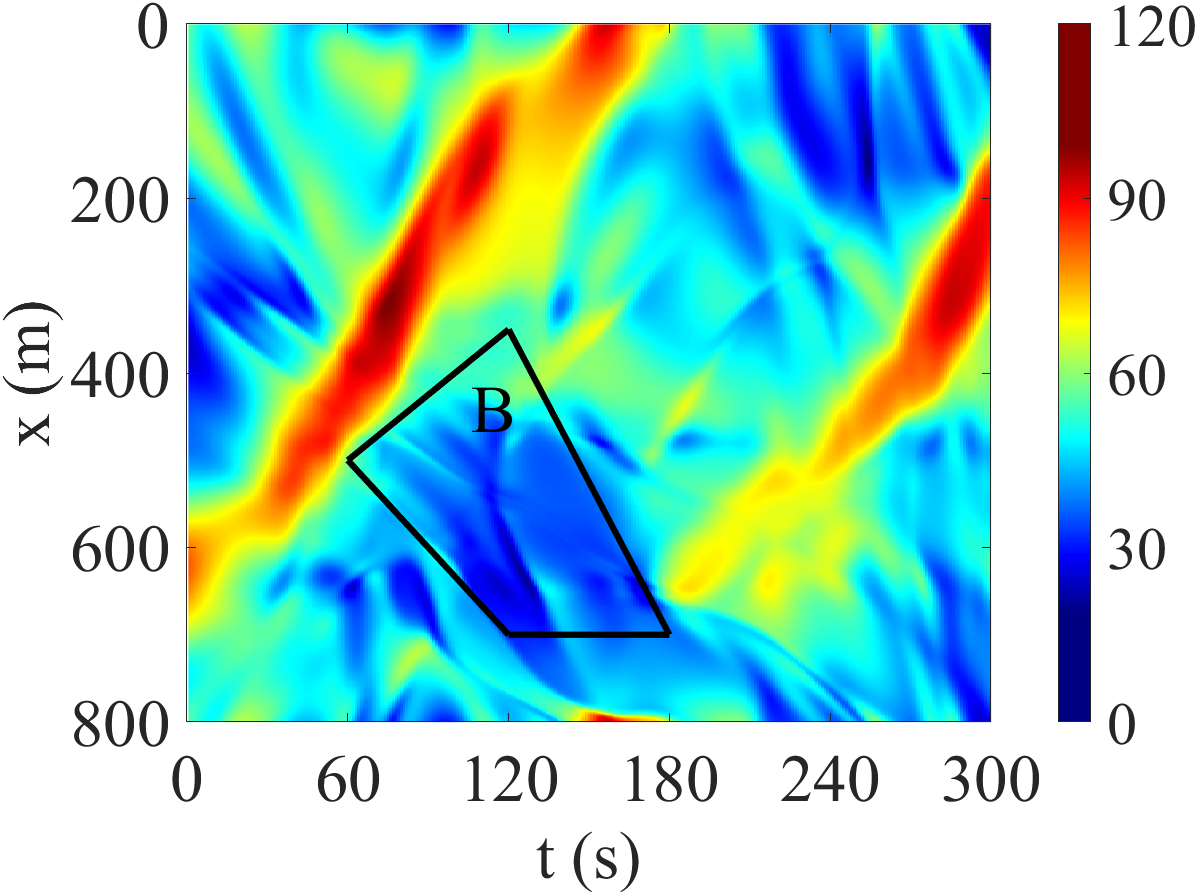}}
    \subfigure[Constant kernel~\eqref{eq:omega constant}]{\includegraphics[width=0.245\linewidth]{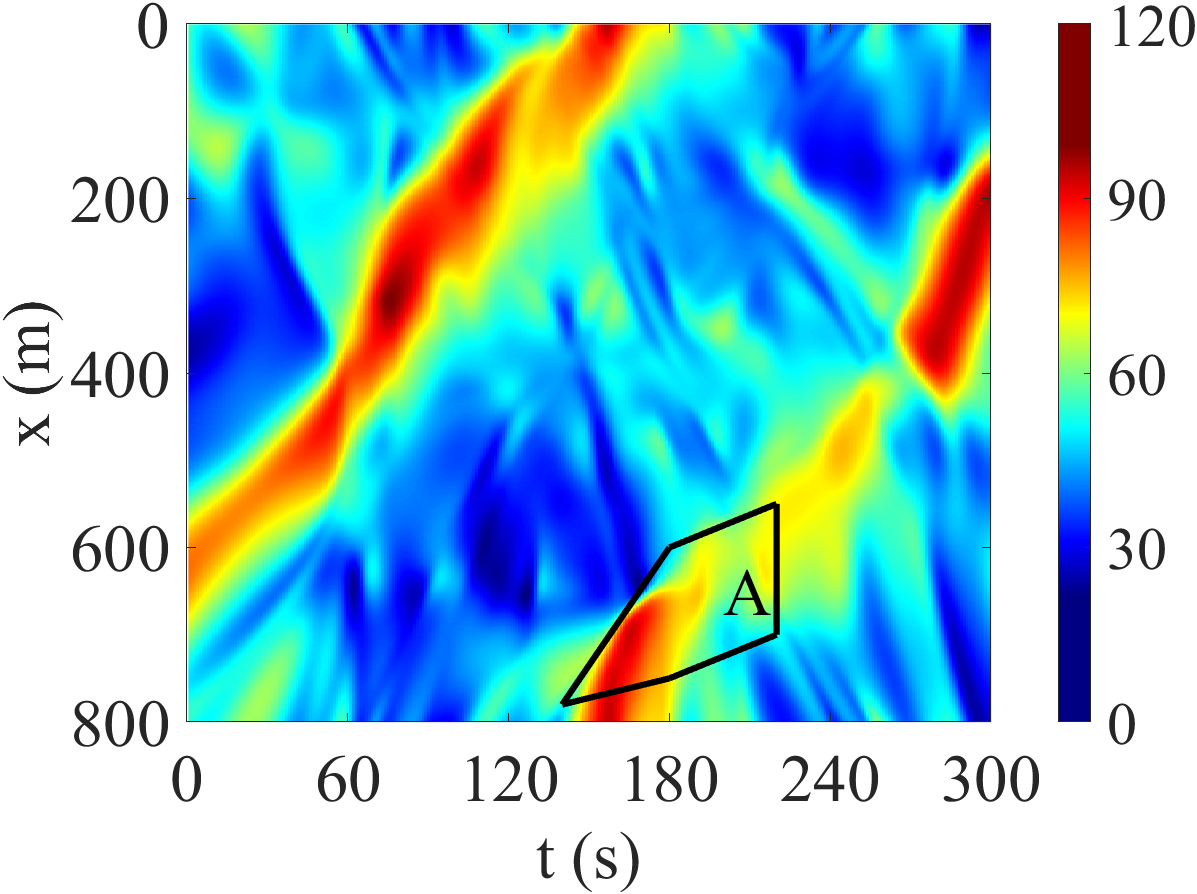}}
    \subfigure[Linear kernel~\eqref{eq:omega decrease}]{\includegraphics[width=0.245\linewidth]{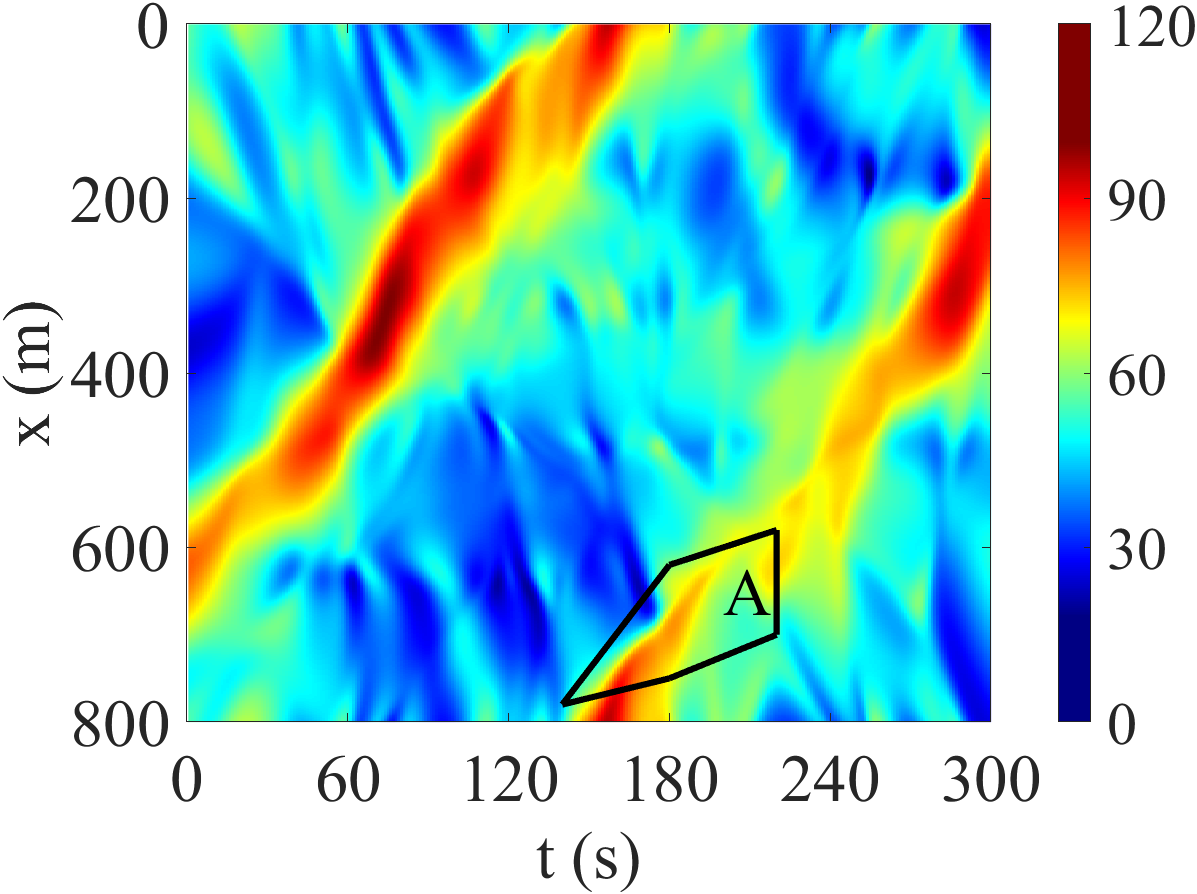}}
    \subfigure[Learned kernel]{\includegraphics[width=0.245\linewidth]{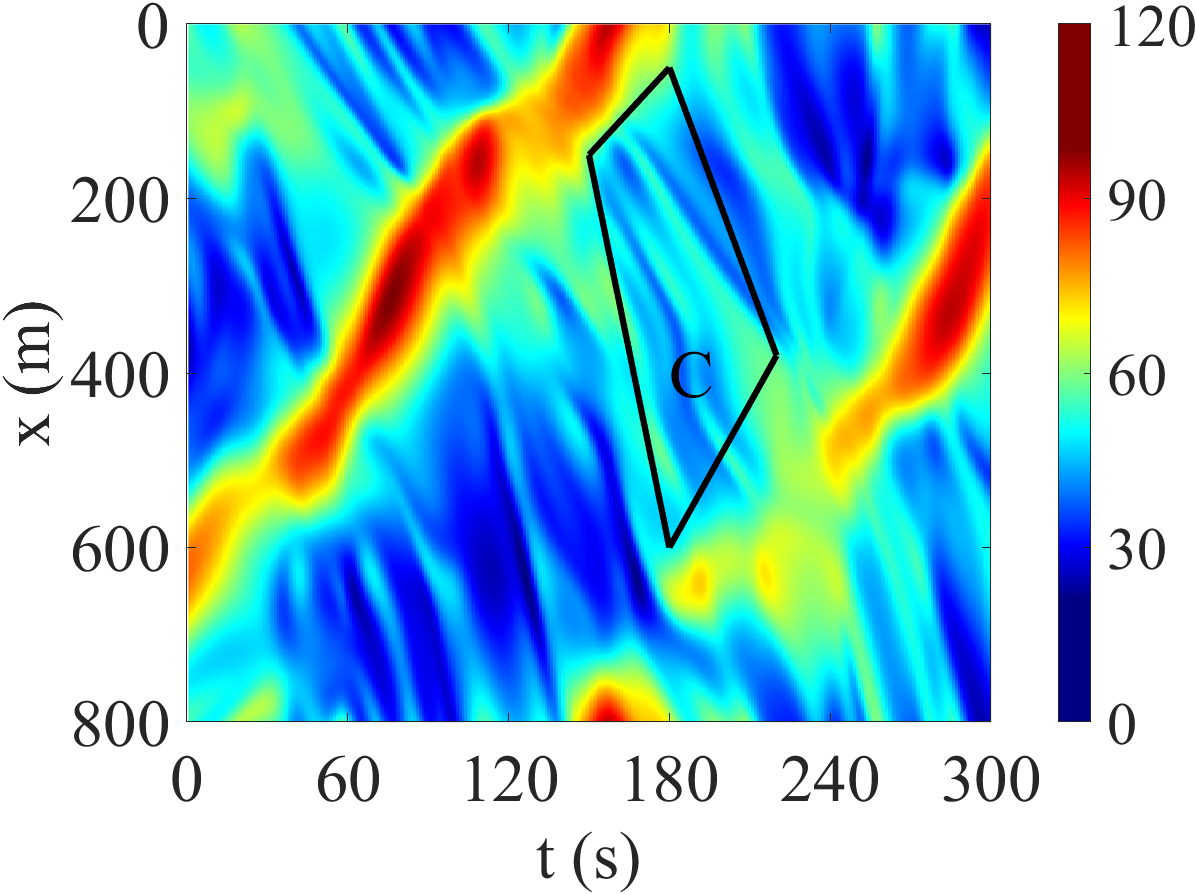}}
    \\
    Underwood~\eqref{eq:FD exp}\\
    \subfigure[Local]{\includegraphics[width=0.245\linewidth]{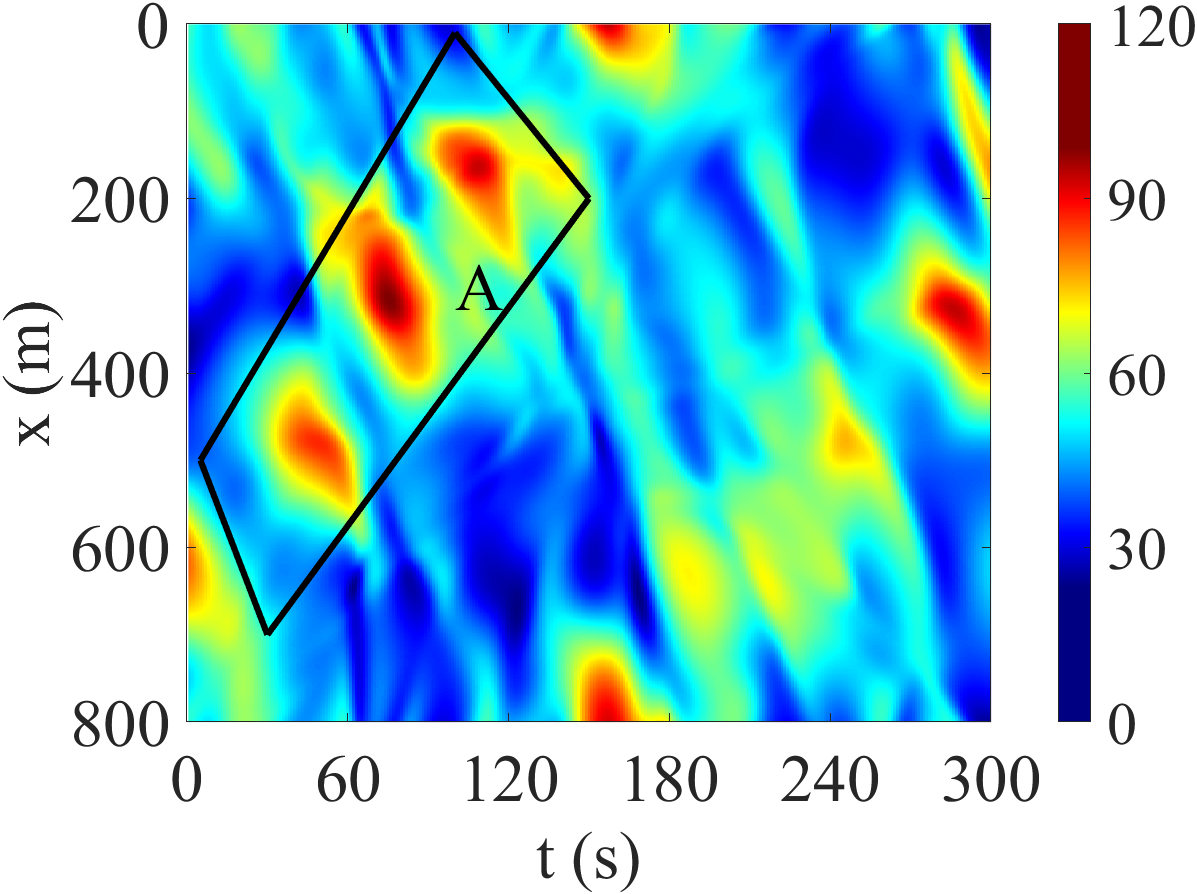}}
    \subfigure[Constant kernel~\eqref{eq:omega constant}]{\includegraphics[width=0.245\linewidth]{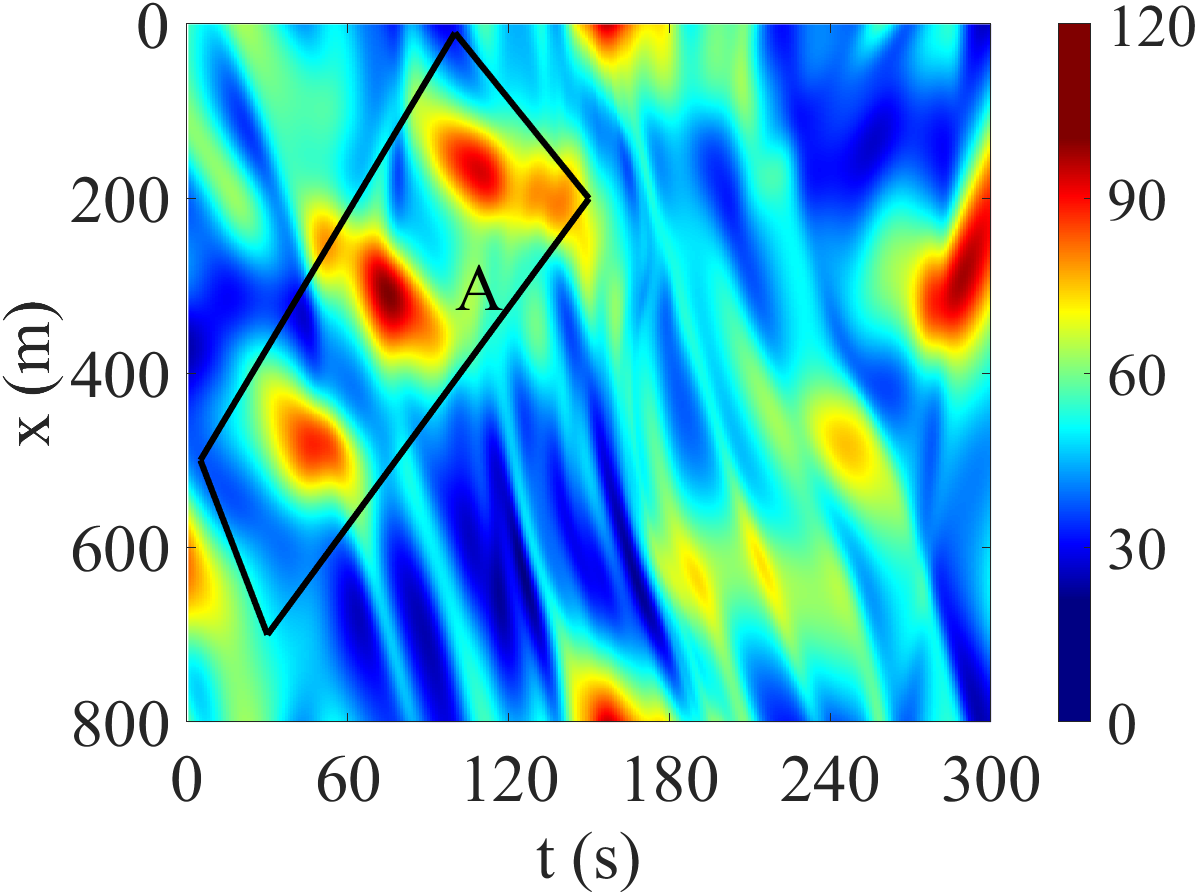}}
    \subfigure[Linear kernel~\eqref{eq:omega decrease}]{\includegraphics[width=0.245\linewidth]{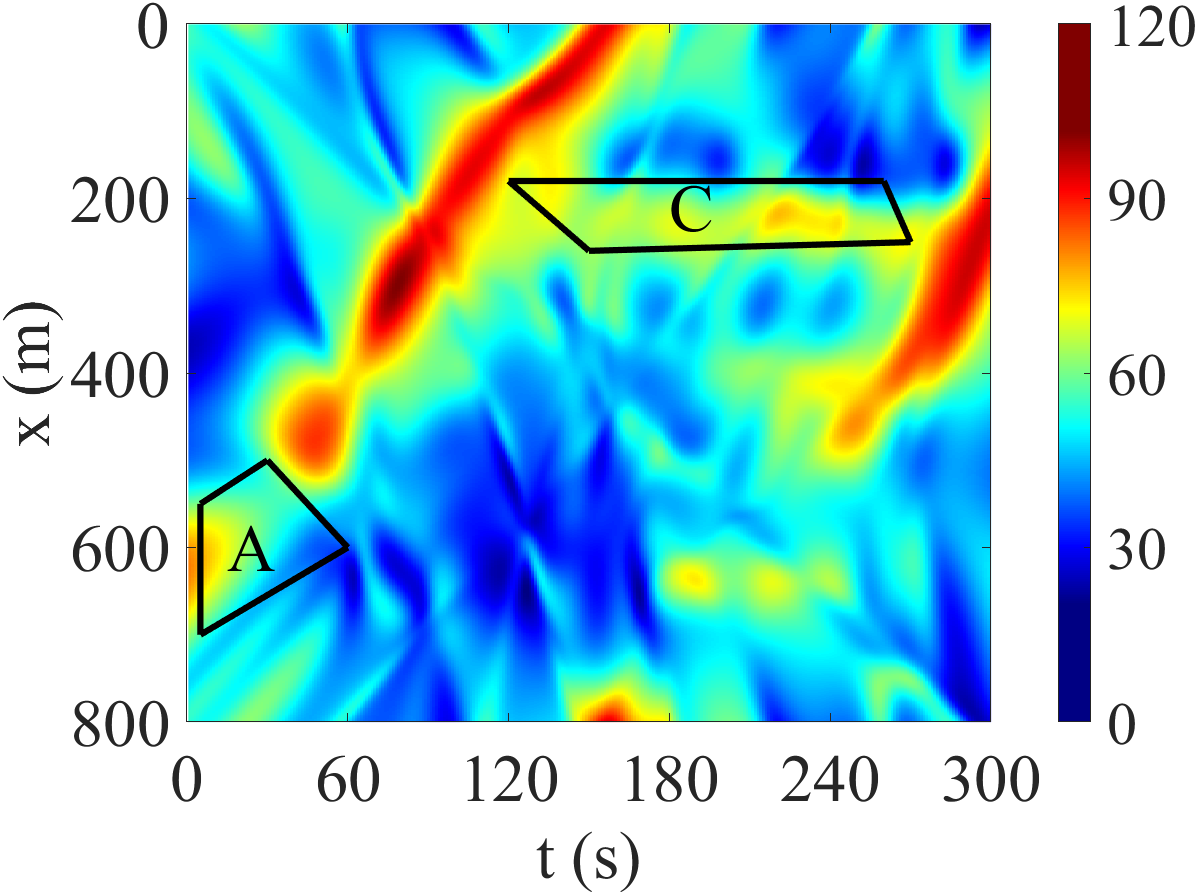}}
    \subfigure[Learned kernel]{\includegraphics[width=0.245\linewidth]{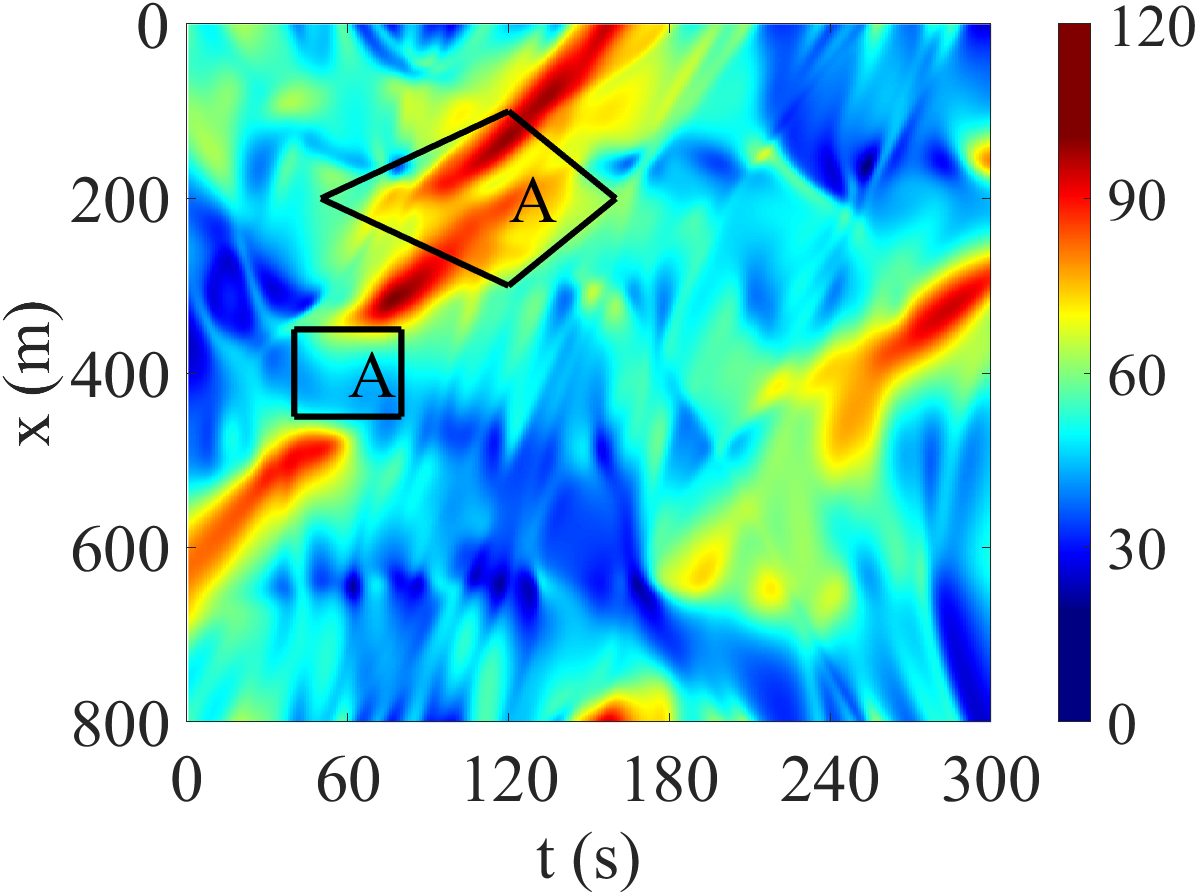}}
    \\
    Drake~\eqref{eq:FD exp2}\\
    \subfigure[Local]{\includegraphics[width=0.245\linewidth]{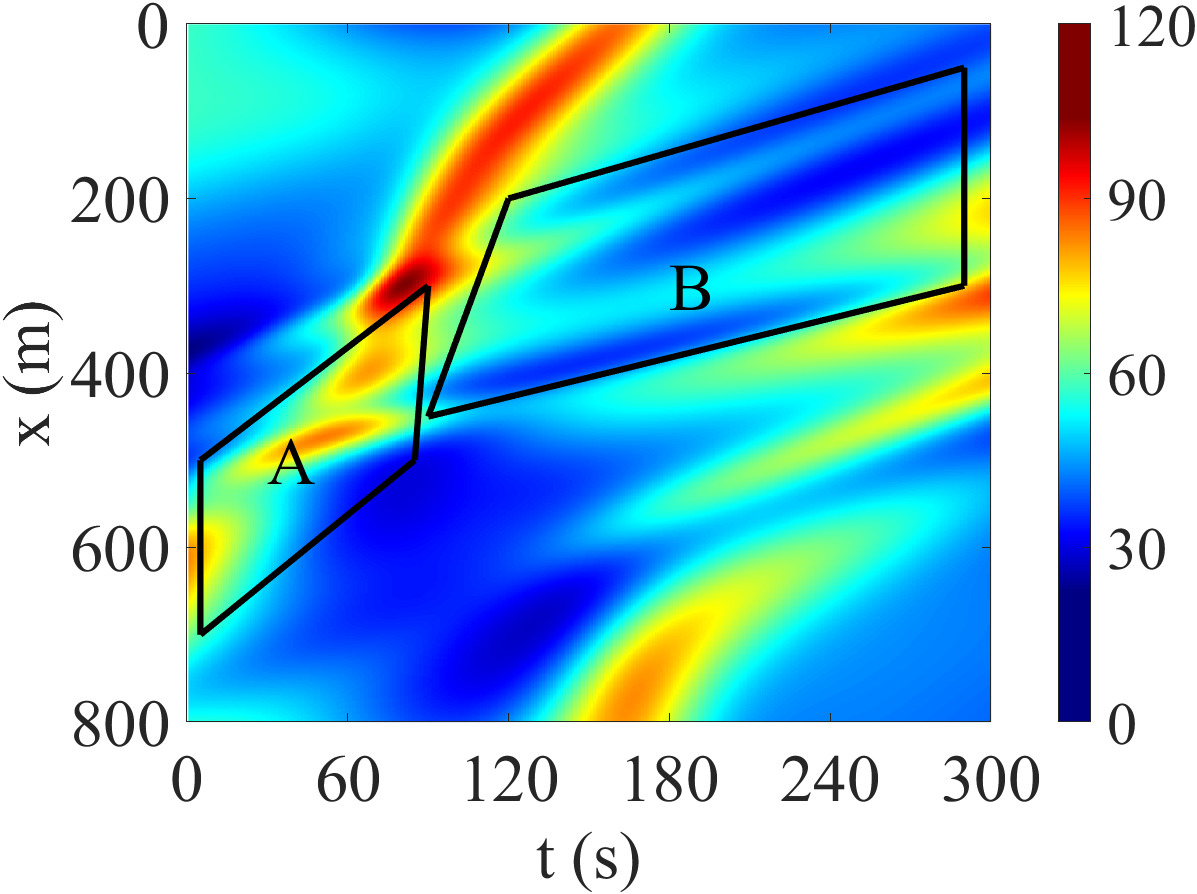}}
    \subfigure[Constant kernel~\eqref{eq:omega constant}]{\includegraphics[width=0.245\linewidth]{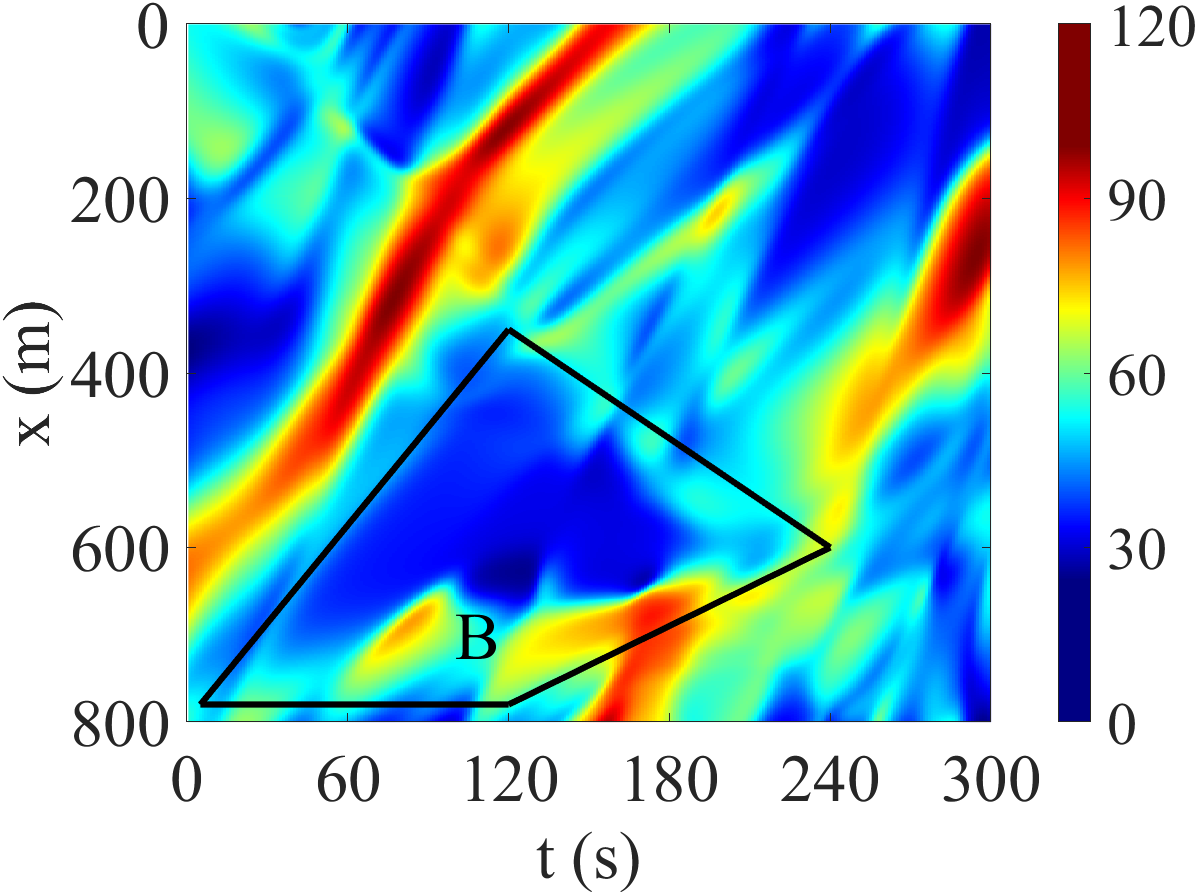}}
    \subfigure[Linear kernel~\eqref{eq:omega decrease}]{\includegraphics[width=0.245\linewidth]{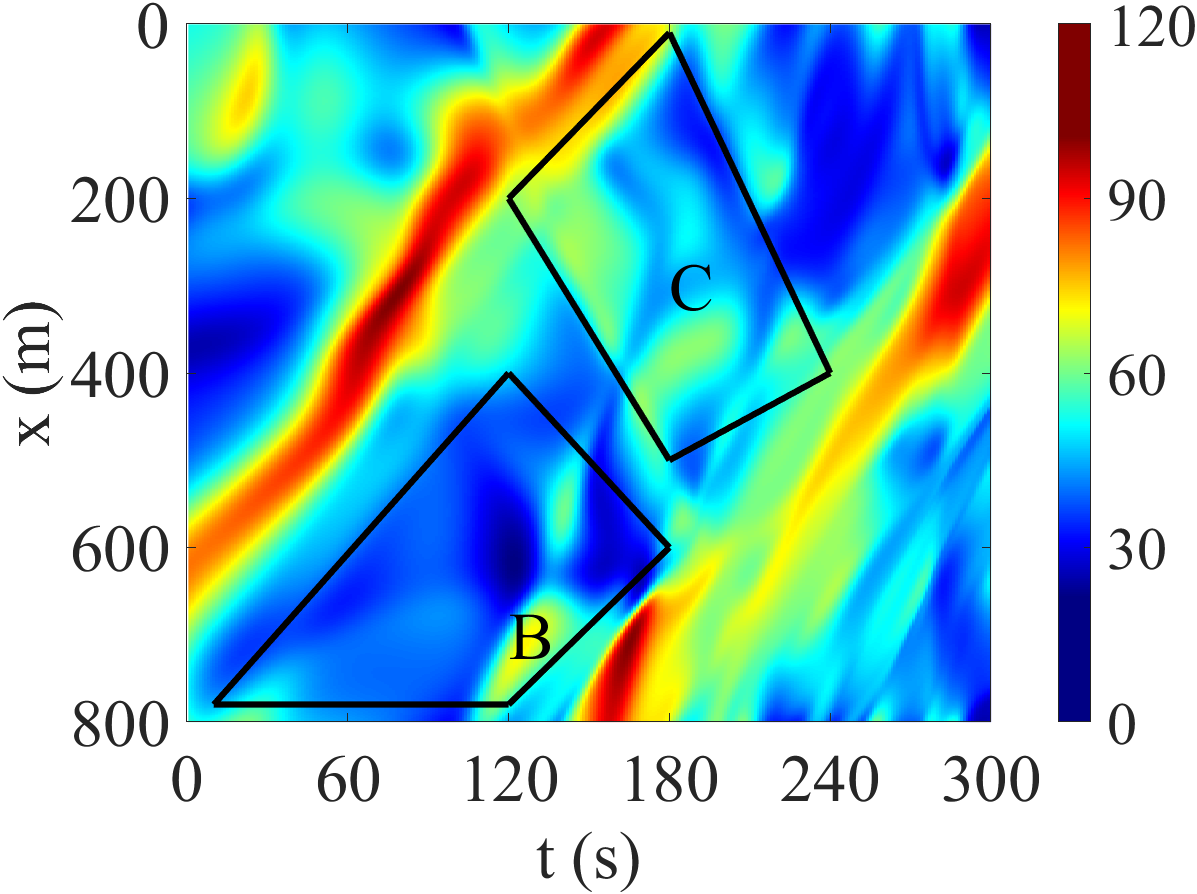}}
    \subfigure[Learned kernel]{\includegraphics[width=0.245\linewidth]{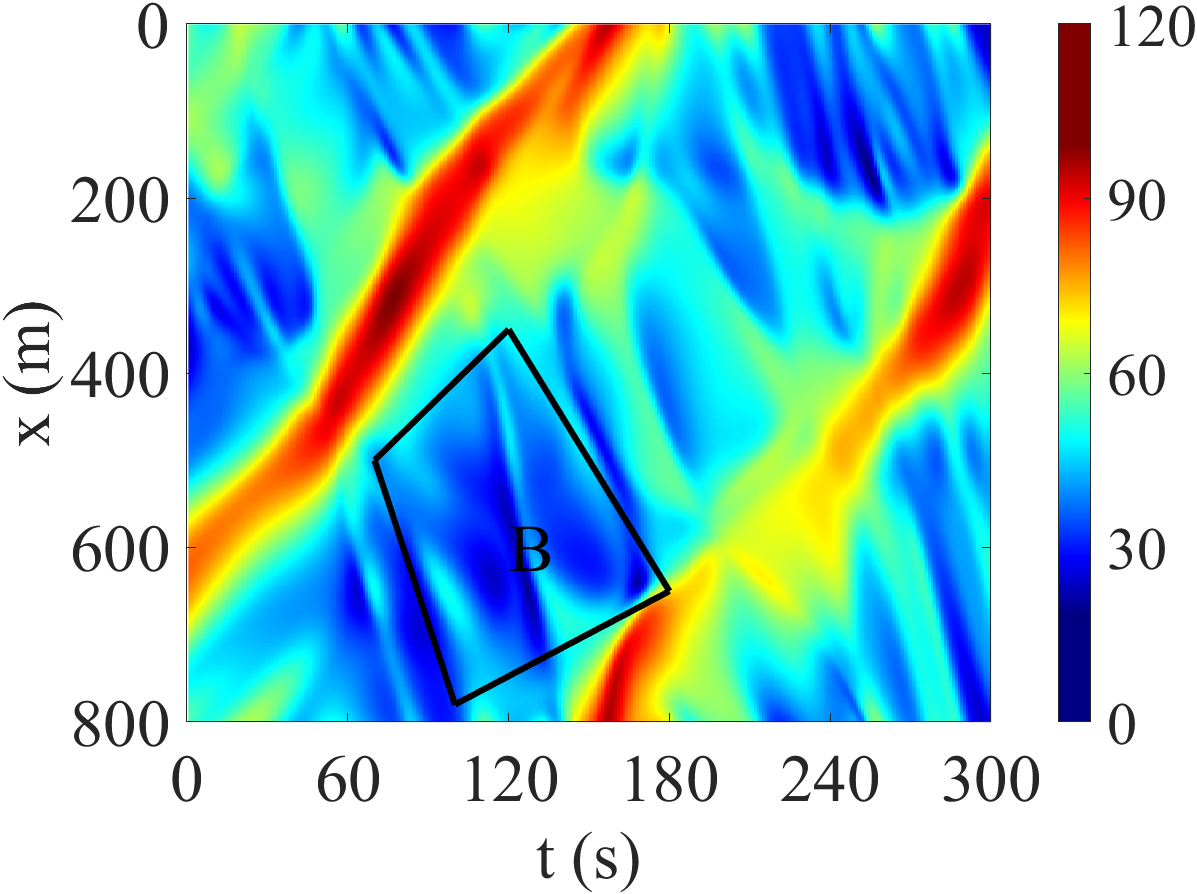}}
    \\
    Learned\\
    \subfigure[Local]{\includegraphics[width=0.245\linewidth]{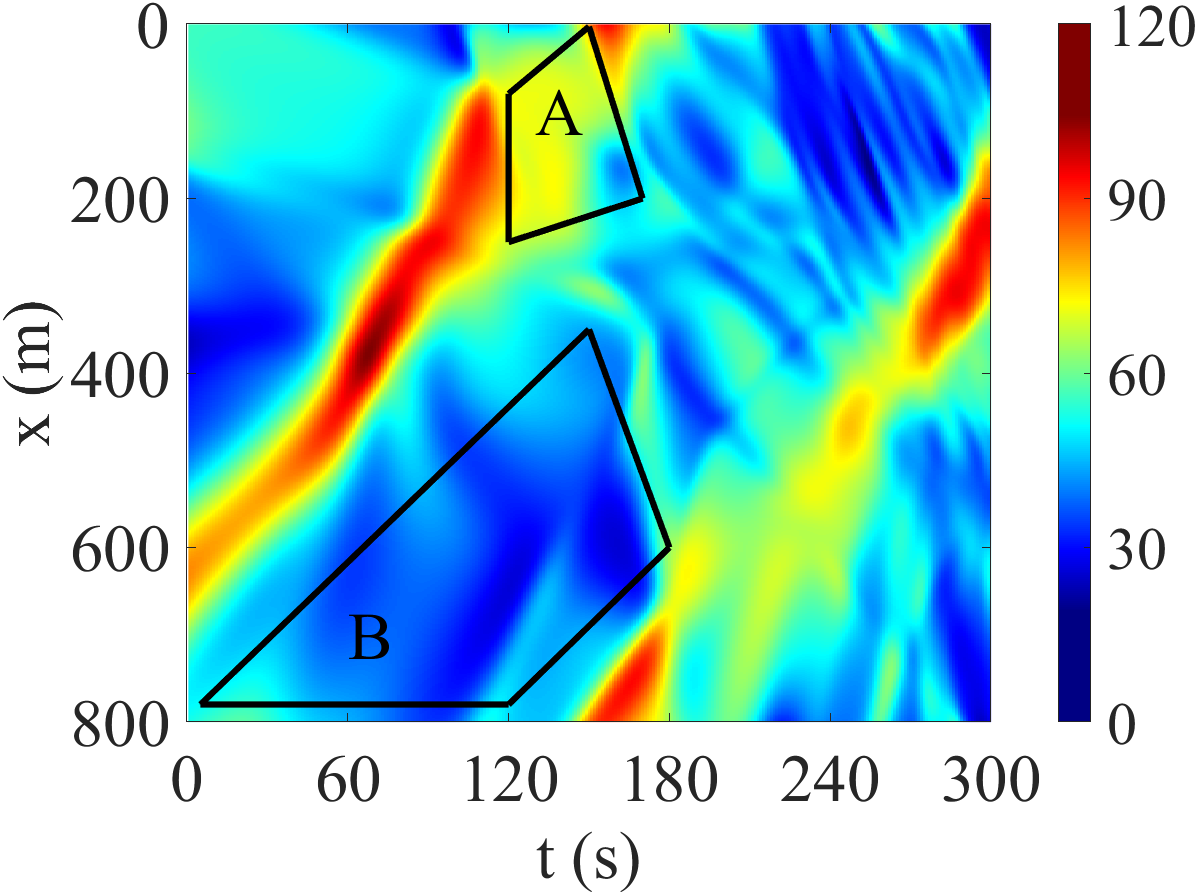}}
    \subfigure[Constant kernel~\eqref{eq:omega constant}]{\includegraphics[width=0.245\linewidth]{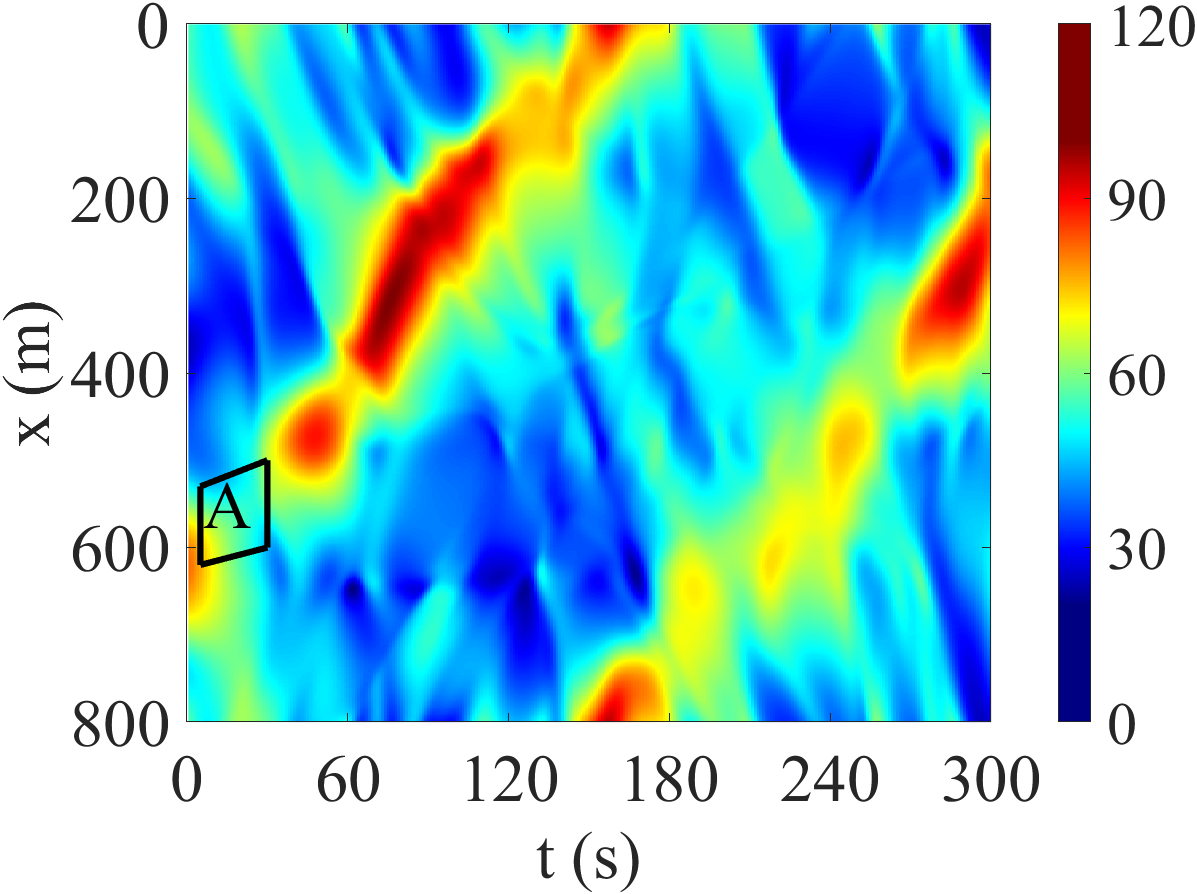}}
    \subfigure[Linear kernel~\eqref{eq:omega decrease}]{\includegraphics[width=0.245\linewidth]{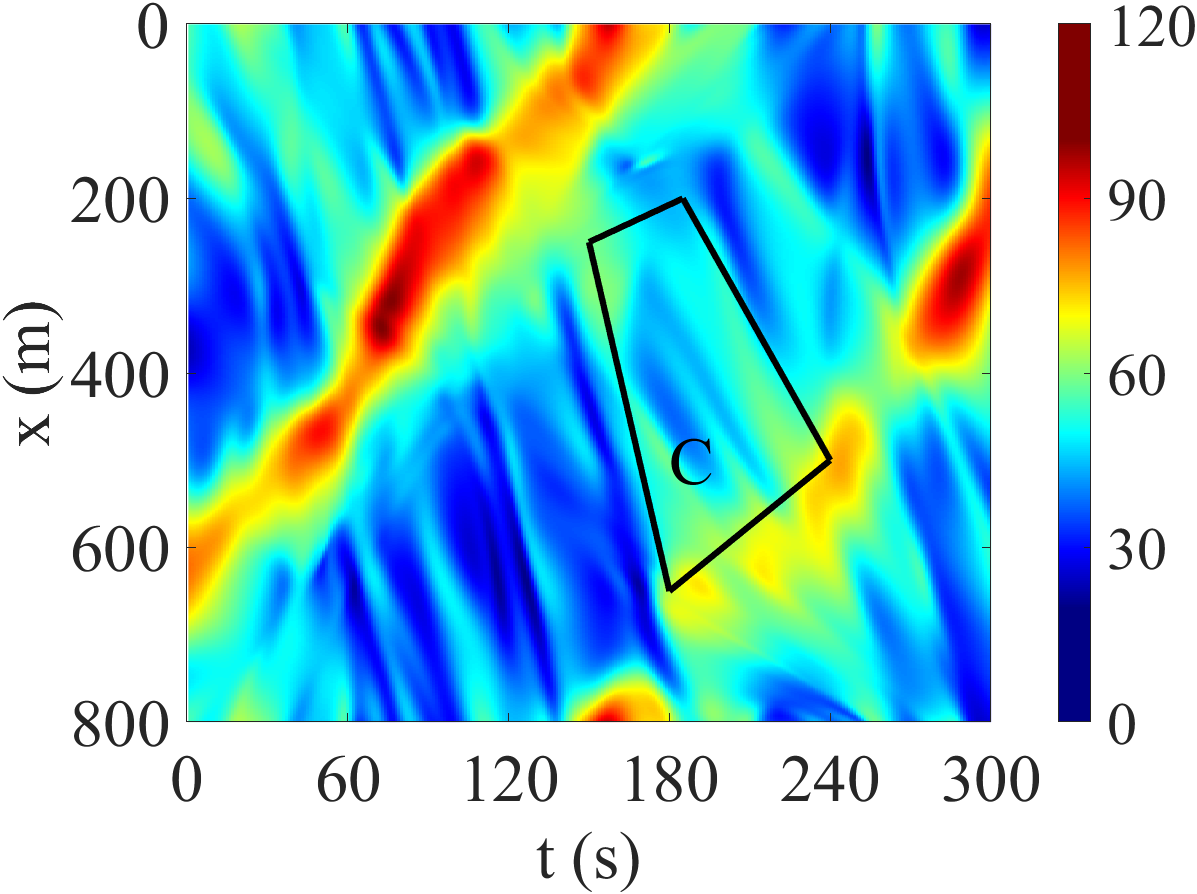}}
    \subfigure[Learned kernel]{\includegraphics[width=0.245\linewidth]{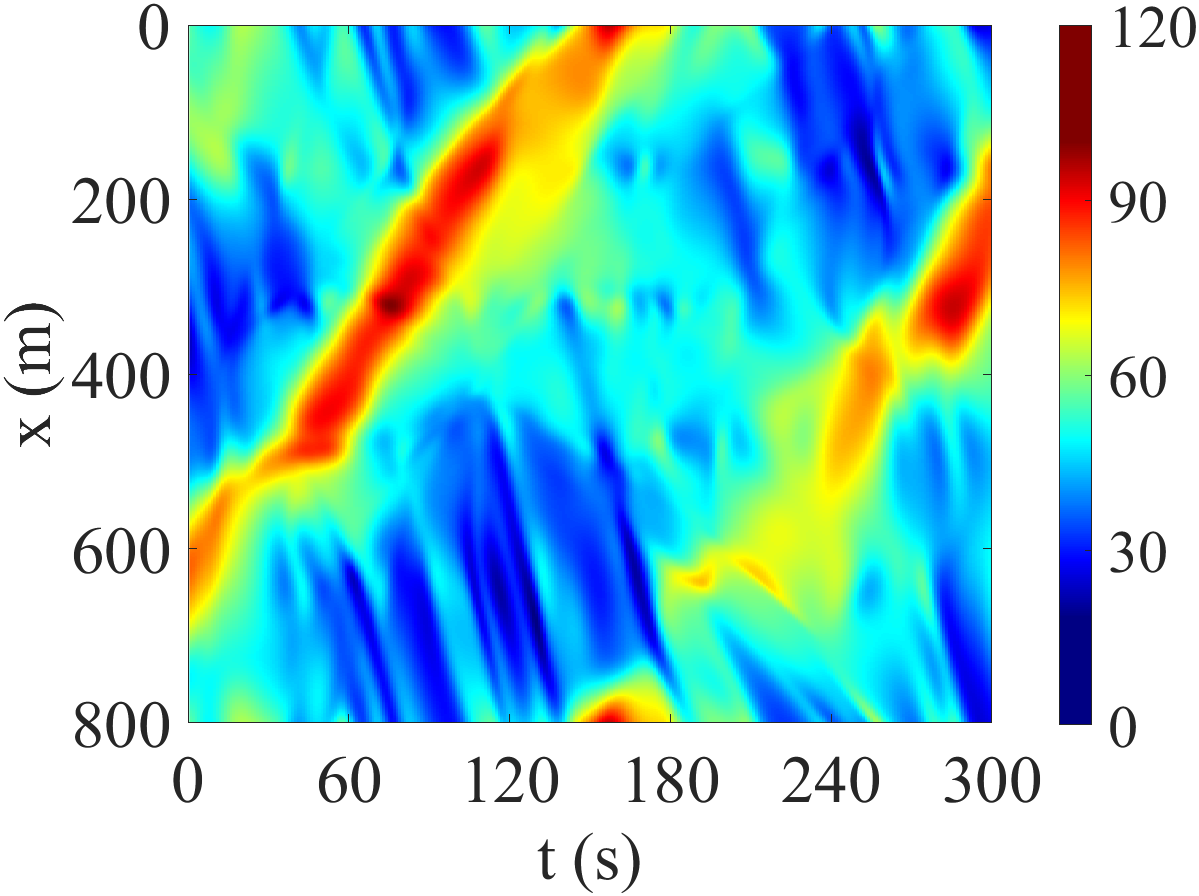}}
    \caption{The ground truth (first row) and learned density dynamics (subfigs (a) to (p)). The text specifies the fundamental diagram, and the subtitle shows the nonlocal kernel. We identify three types of traffic dynamics from the ground truth, and label inaccurate estimation in the learned dynamics.}
    \label{fig:rho dynamics}
\end{figure}

\subsection{Training settings}

For the variable $\theta$, we use a fully connected feedforward neural network with six hidden layers and 64 neurons in each hidden layer. 
For the fundamental diagram $\theta_v$, we use a fully connected feedforward neural network with two hidden layers and 64 neurons in each hidden layer. 

We use $N_l = 5$ loop detectors evenly distributed among the ring road including one at the boundary $x=0$. 
We train NN parameters  by first running 50,000 iterations of ADAM for rough training and then using L-BFGS to refine the NN. We randomly choose $N_P = 512$ points from $\mathcal{G}$ to evaluate the physics dynamic loss $L_{\mathrm{phy-d}}$. We decide coefficients $\alpha_{i}$ in data loss $L_{\mathrm{data}}$ by grid search. We set the penalty coefficients as $p_{v,1} = p_{v,2} p_{\omega,1} = p_{\omega,2}=10^4$.

\section{Analyze nonlocal effect on ring-road traffic }

In this section, we analyze the learned dynamics with local and nonlocal LWR PDEs from two aspects: traffic dynamics in Section~\ref{sec:analyze dynamics} and static speed-density fundamental diagram in Section~\ref{sec:analyze FD}.  

\subsection{Nonlocal effect on traffic dynamics}\label{sec:analyze dynamics}

{\begin{table}\small 
    \centering
    \caption{Estimation error (\%)  of density}
    \label{tab:error rho}
    \begin{tabular}{c|c|c|c|c}
    \hline
    \diagbox{FD}{Kernel}  & Local & Constant~\eqref{eq:omega constant} & Linear~\eqref{eq:omega decrease} & Learned \\ \hline
    Greenshields~\eqref{eq:FD greenshields} & 16.80 & 16.07 & 14.08 & 12.70 \\ \hline
    Underwood~\eqref{eq:FD exp} & 19.30 & 20.36 & 17.50  &  17.27 \\ \hline
    Drake~\eqref{eq:FD exp2} & 18.71 & 20.62 & 17.82 & 14.66 \\ \hline
    Learned & 14.92 & 12.70  &  12.20 & 11.90  \\ \hline
    \end{tabular}
\end{table}}

\begin{figure}
    \centering
    \subfigure[$\eta=35$ m]{\includegraphics[width=0.245\linewidth]{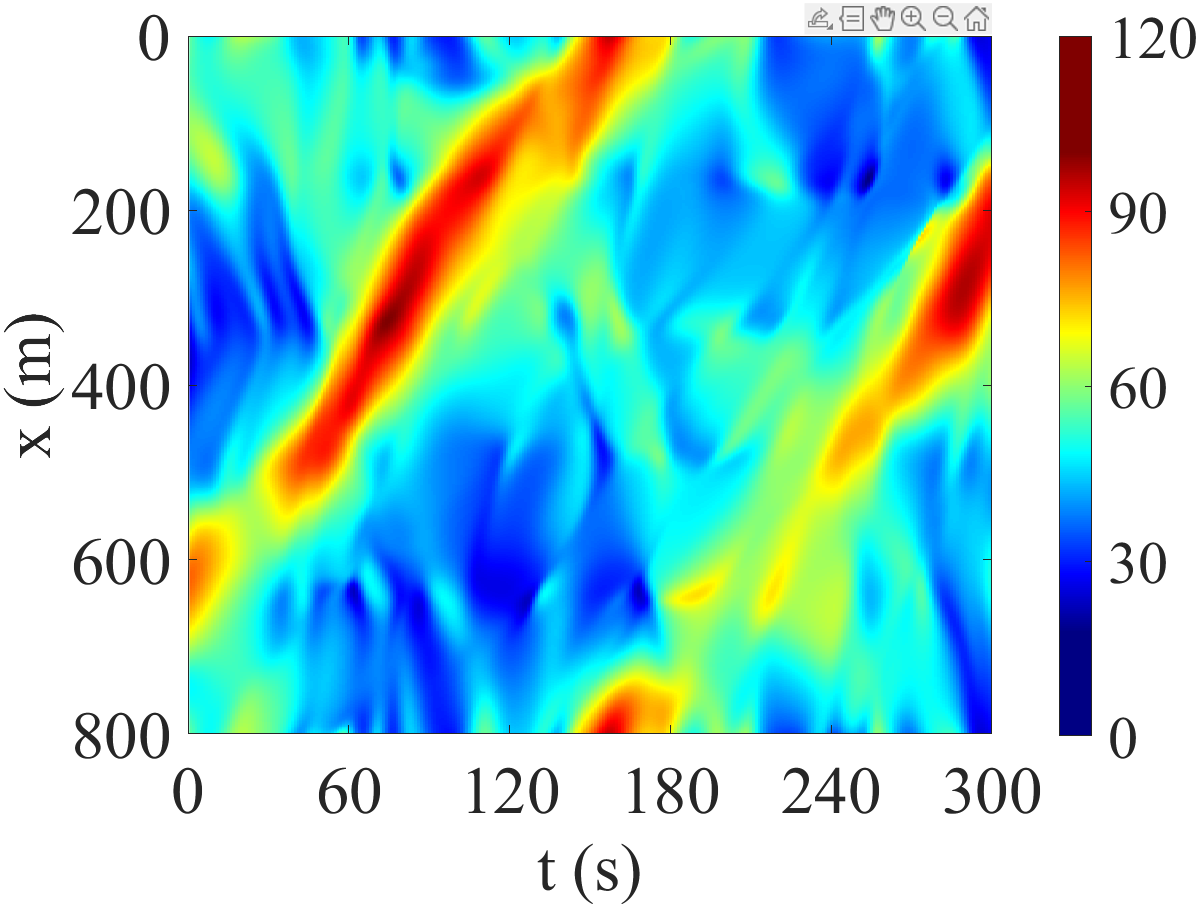}}
    \subfigure[$\eta=40$ m]{\includegraphics[width=0.245\linewidth]{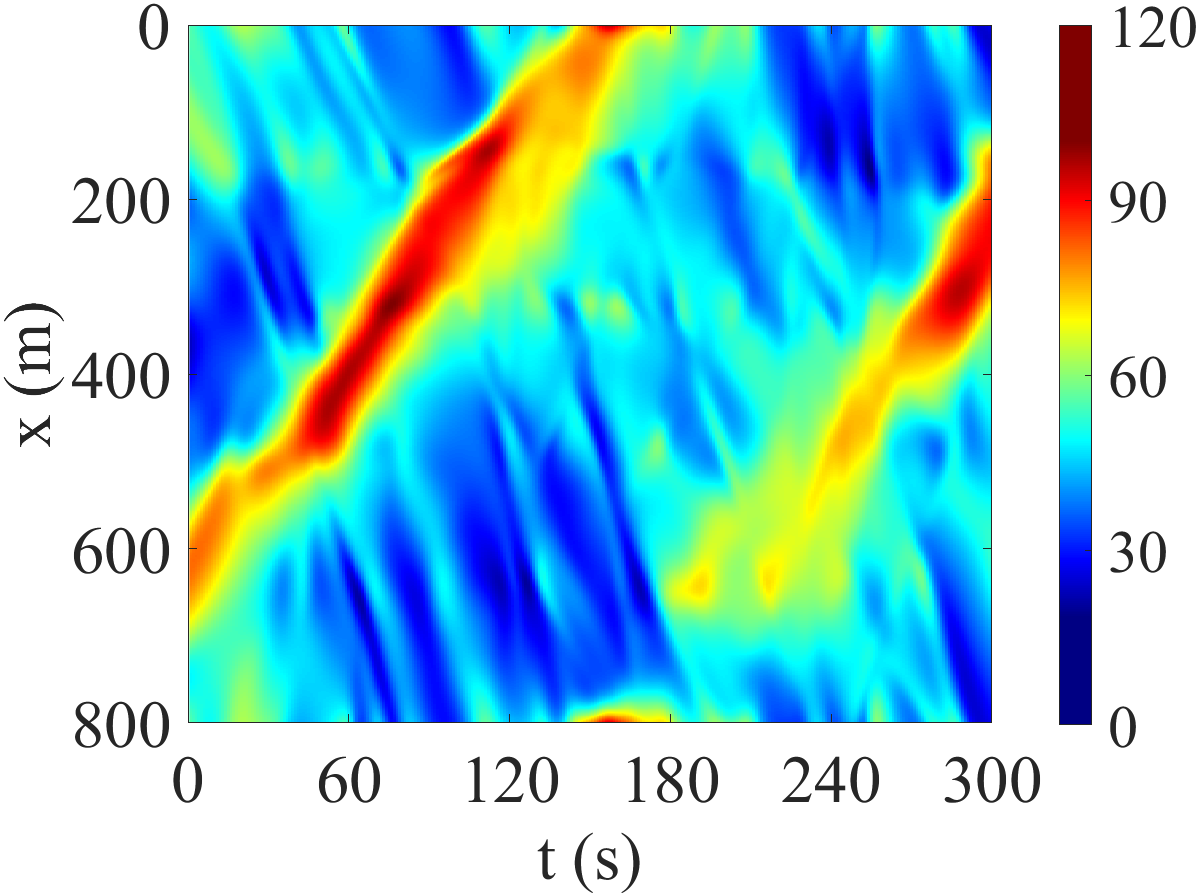}}
    \subfigure[$\eta=45$ m]{\includegraphics[width=0.245\linewidth]{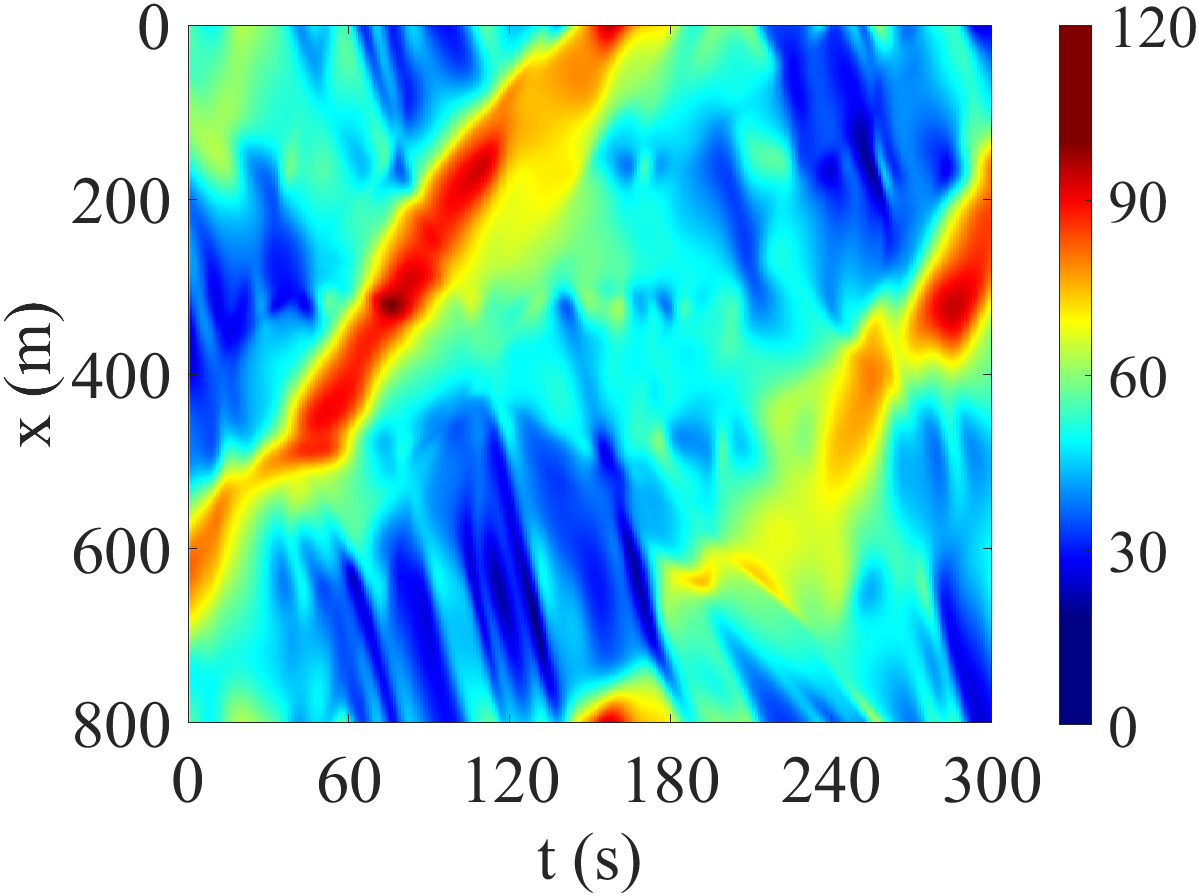}}
    \subfigure[$\eta=50$ m]{\includegraphics[width=0.245\linewidth]{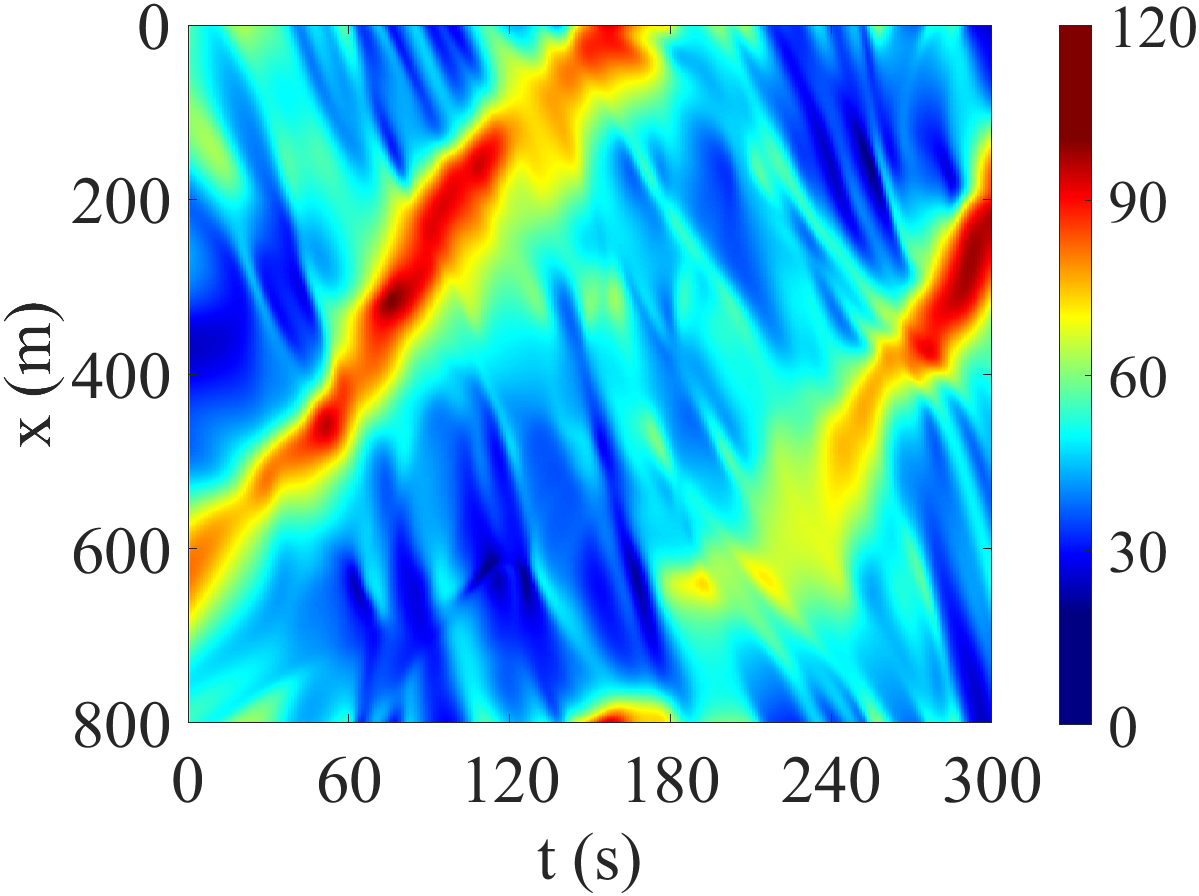}}
    \vspace{-2em}
    \caption{The learned density dynamics with varying kernel length $\eta$.}
    \label{fig:compare eta dynamics}
    \vspace{-1em}
\end{figure}

To evaluate the error between learned density $\hat \rho$ and actual density, we use relative RMSE:
\begin{equation}
    \setlength{\abovedisplayskip}{\abovegap}
    \setlength{\belowdisplayskip}{\belowgap}
    \mathrm{E}_{\rho} = \frac{\sqrt{\sum_{i,j} \left( \hat  \rho (t_i,x_j) - \rho(t_i,x_j) \right)^2  }}{\sqrt{\sum_{i,j} \left( \rho (t_i,x_j) \right)^2}} \times 100\%.
\end{equation}
In Table~\ref{tab:error rho}, we give the estimation error with different fundamental diagrams and look-ahead kernels. From each row, we see that given a fundamental diagram, such as  Underwood and Drake,  using constant look-head kernel~\eqref{eq:omega constant} may instead increase the estimation error. Using NN to learn a look-ahead kernel reduces the estimation error for all four cases of fundamental diagrams.  Comparing the estimation error of the same look-ahead kernel and different fundamental diagrams in each column, we find that learning a fundamental diagram via NN consistently has lower estimation errors.  

In Fig.~\ref{fig:rho dynamics}, we plot the learned density dynamics. In each row, we see that with the learned look-ahead kernel, there is a more accurate traffic pattern than both the local LWR  and nonlocal LWR with linear and constant kernels. Taking the last row as an example, we see that local LWR fails to learn the propagation of free traffic wave as Fig.~\ref{fig:rho dynamics}(m) shows, and nonlocal LWR with a constant look-ahead kernel causes false propagation of free flow as Fig.~\ref{fig:rho dynamics}(n) shows. This not only validates the existence of nonlocal effect in real traffic but also underscores the importance of choosing physics prior for PIDL.

Comparing the estimation obtained by different fundamental diagrams, we see that the learned fundamental diagram yields more accurate dynamics. For example,  Fig.~\ref{fig:rho dynamics}(d) shows that the Greenshields fundamental diagram estimates a distorted propagation of the stop-and-go oscillation. For the Underwood fundamental diagram, as the black box in Fig~\ref{fig:rho dynamics}(h) shows, it fails to learn the congestion wave and gets false free traffic flow.

\begin{figure}
    \centering
    \subfigure[$\eta=35$ m]{\includegraphics[width=0.245\linewidth]{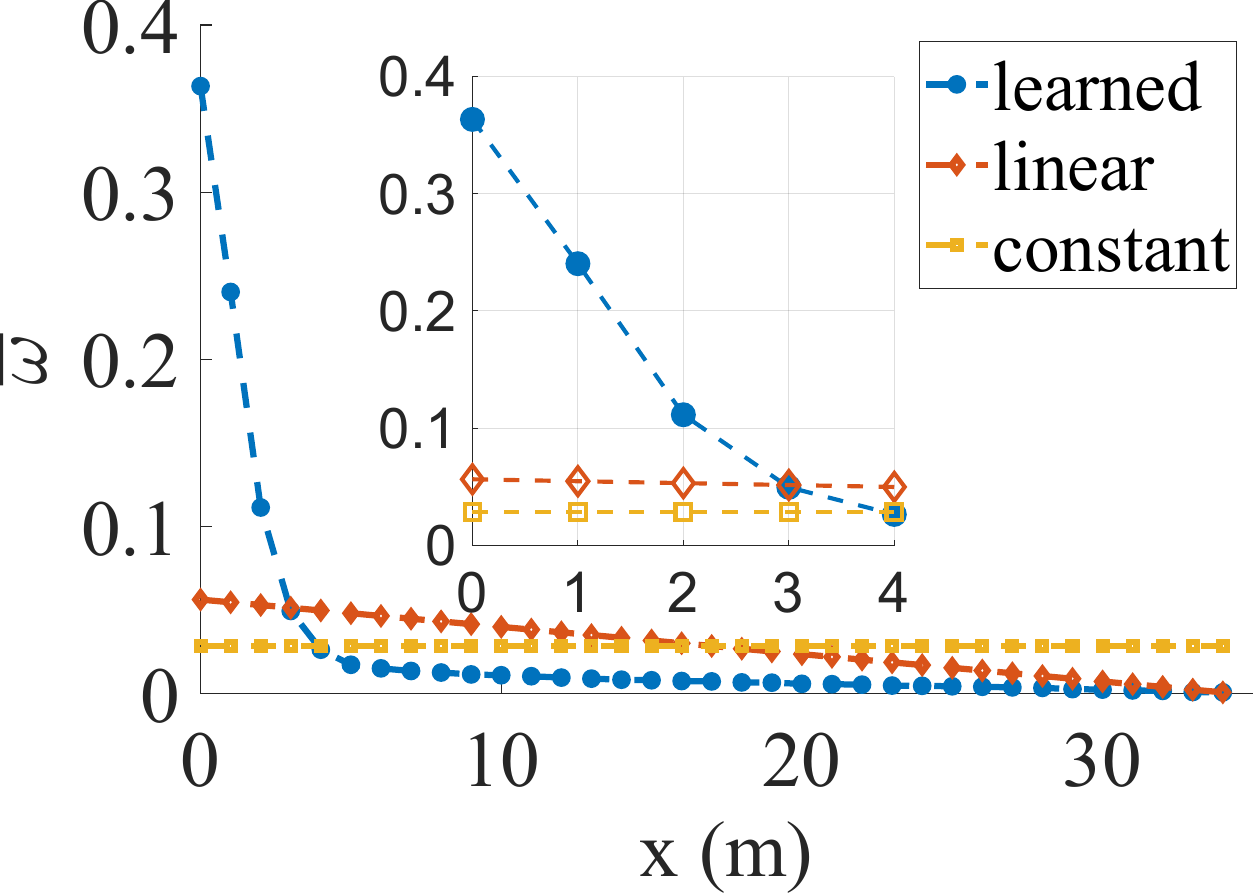}}
    \subfigure[$\eta=40$ m]{\includegraphics[width=0.245\linewidth]{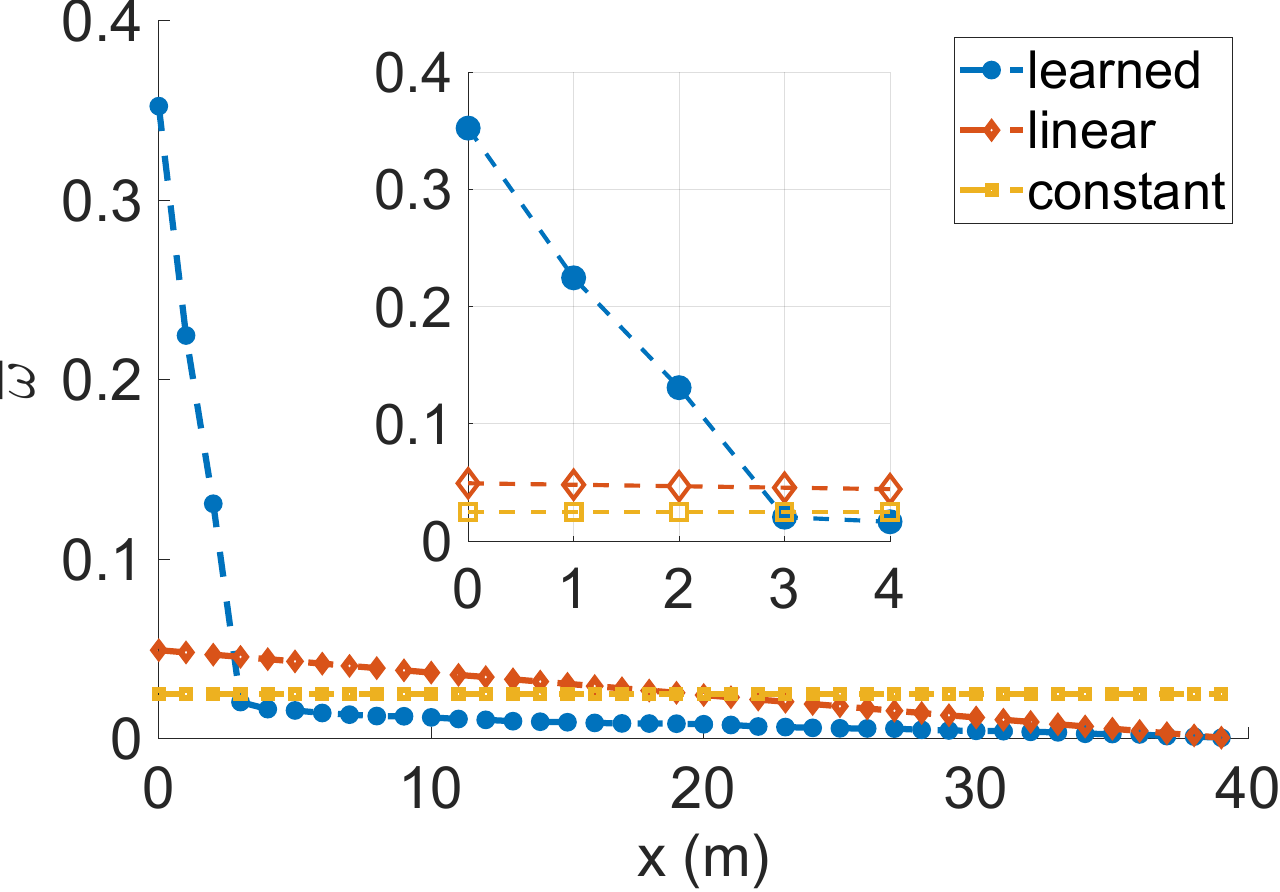}}
    \subfigure[$\eta=45$ m]{\includegraphics[width=0.245\linewidth]{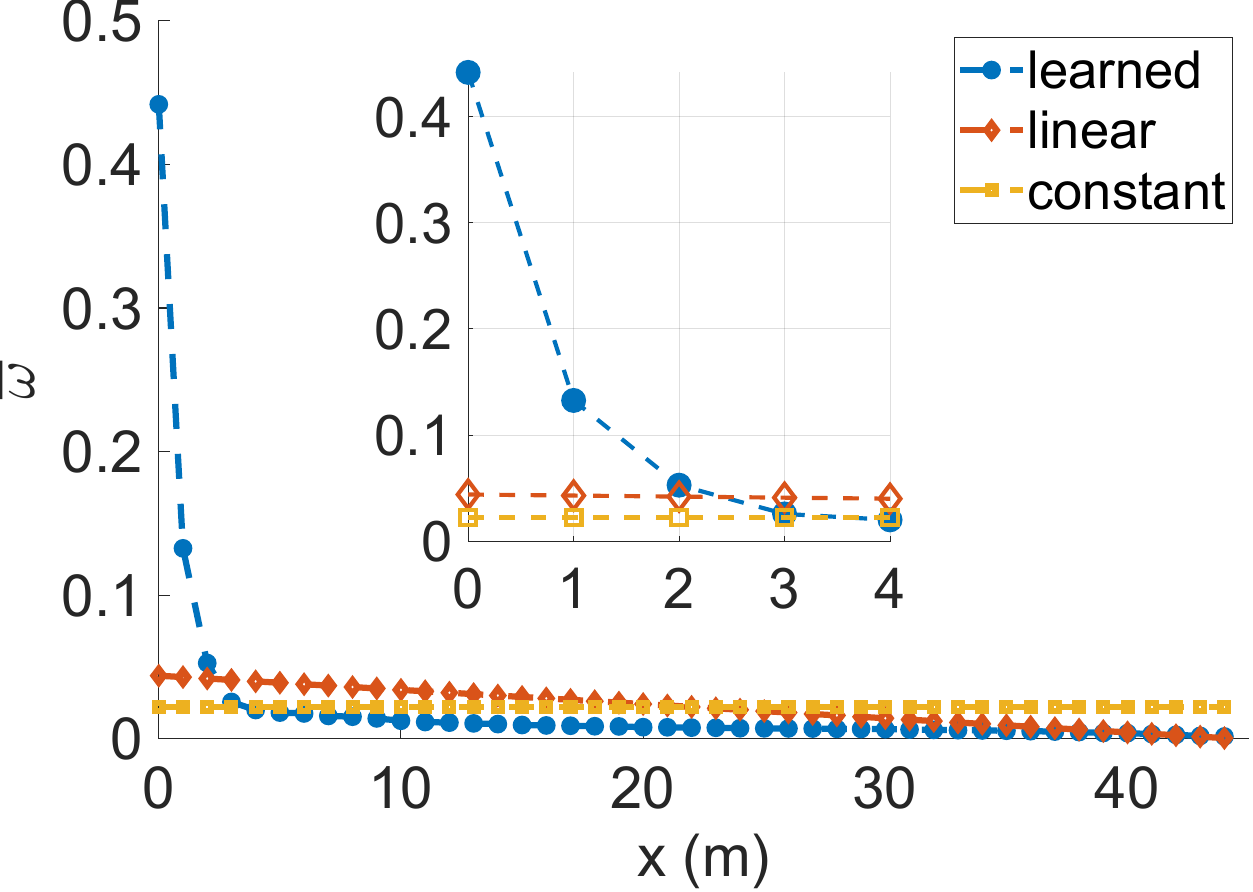}}
    \subfigure[$\eta=50$ m]{\includegraphics[width=0.245\linewidth]{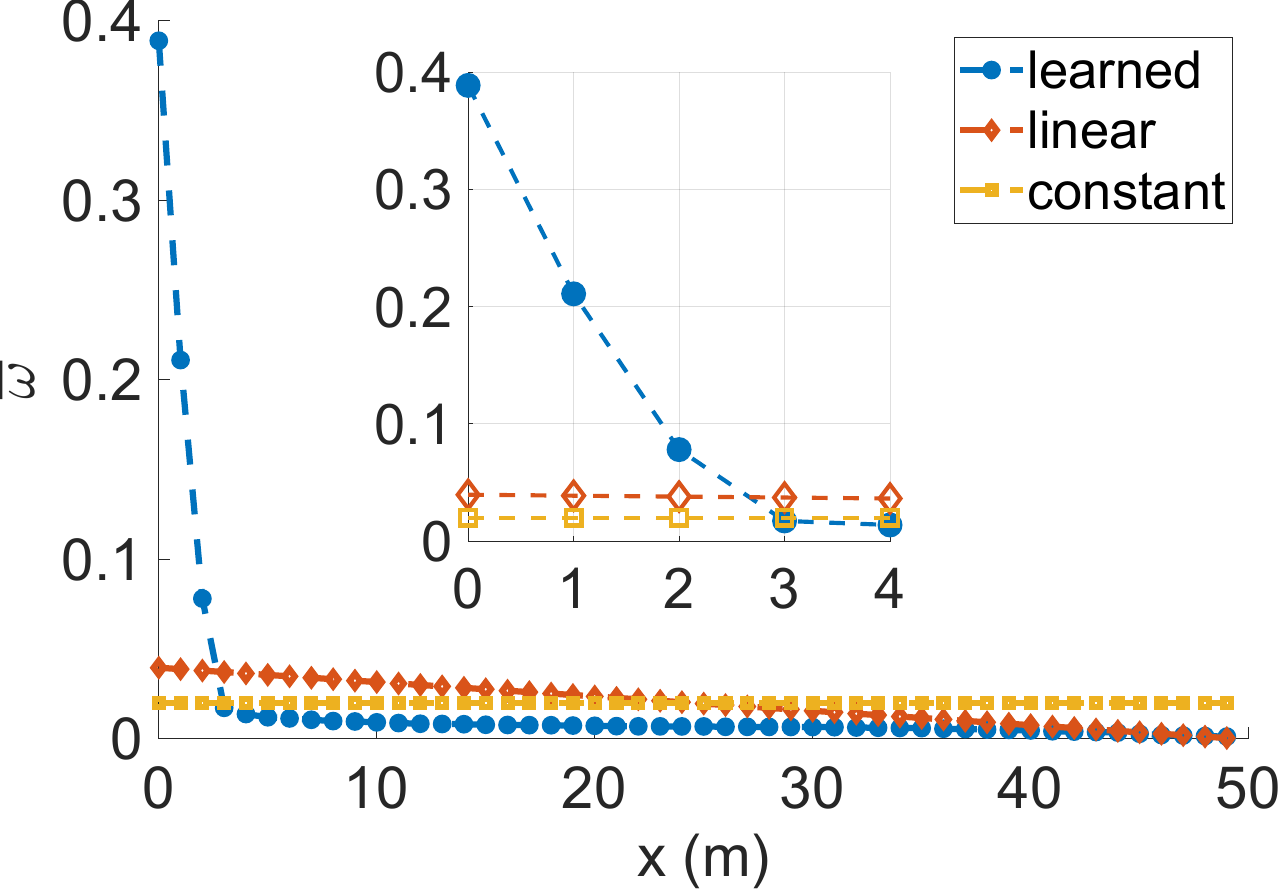}}
    \vspace{-2em}
    \caption{The kernel function learned by NN with varying kernel length $\eta$.}
    \label{fig:compare eta omega}
\end{figure}

We run simulations with the kernel length $\eta$ ranging from 5 m to 60 m, and find that 35 m to 50 m yields more accurate estimated dynamics. In Fig.~\ref{fig:compare eta dynamics}, we give the learned density with varying kernel length. We see that all of them have an accurate estimation of traffic waves. 
In Fig.~\ref{fig:compare eta omega}, we plot the learned kernel with different kernel length $\eta$. We see that the optimal kernel length has a similar pattern: a  majority of the nonlocal effect is within 5 to 10 meters, which is approximately the length of one to two vehicles. For the kernel shown in the subfigures in Fig.~\ref{fig:compare eta omega}, the look-ahead kernel within  4 meters accounts for $68 \% - 80 \%$ of the whole look-ahead effect. This indicates that the speed is mainly dictated by the vehicle and its leader vehicle.

\begin{figure}
    \centering
    \subfigure[Local]{\includegraphics[width=0.245\linewidth]{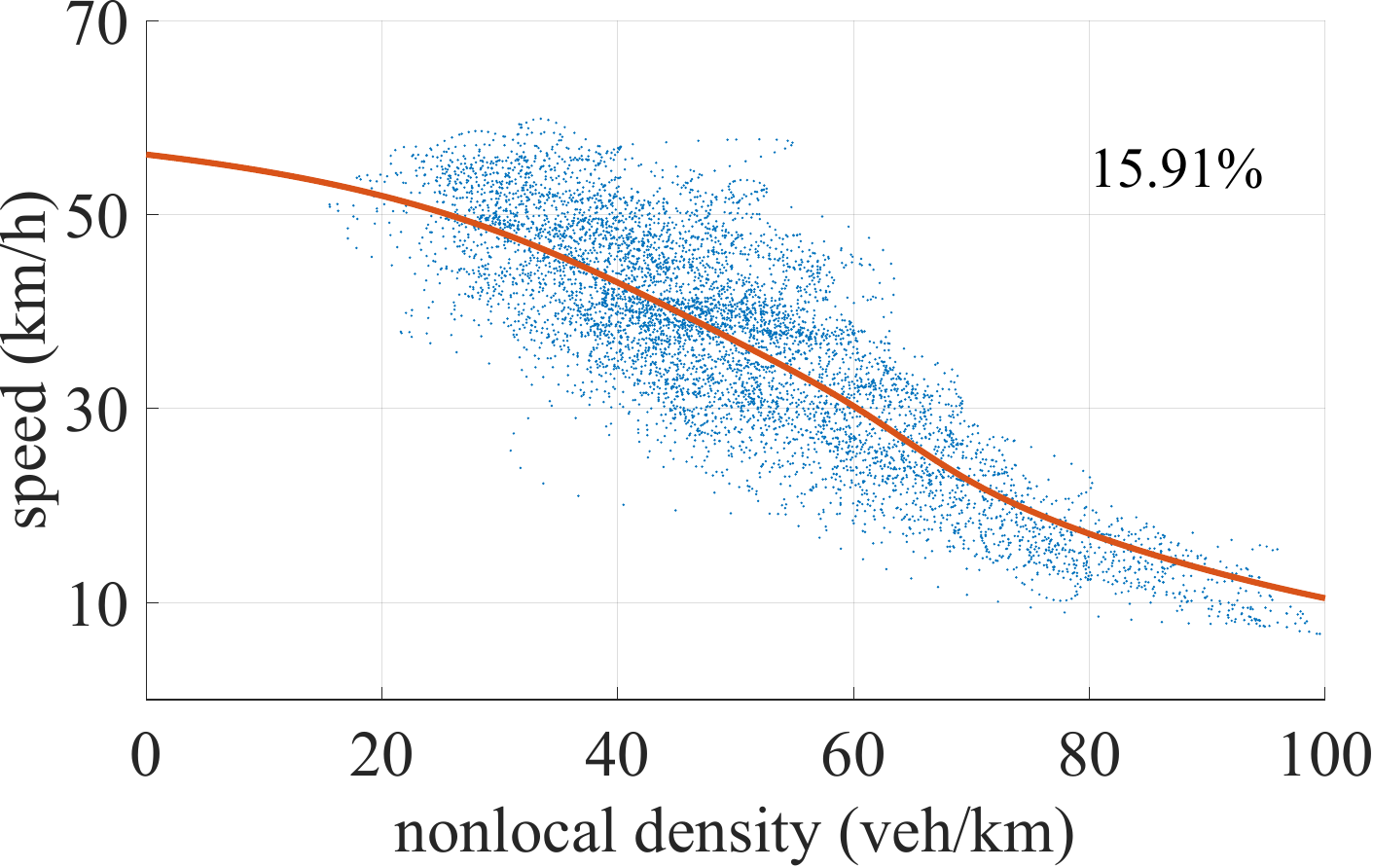}}
    \subfigure[Constant kernel~\eqref{eq:omega constant}]{\includegraphics[width=0.245\linewidth]{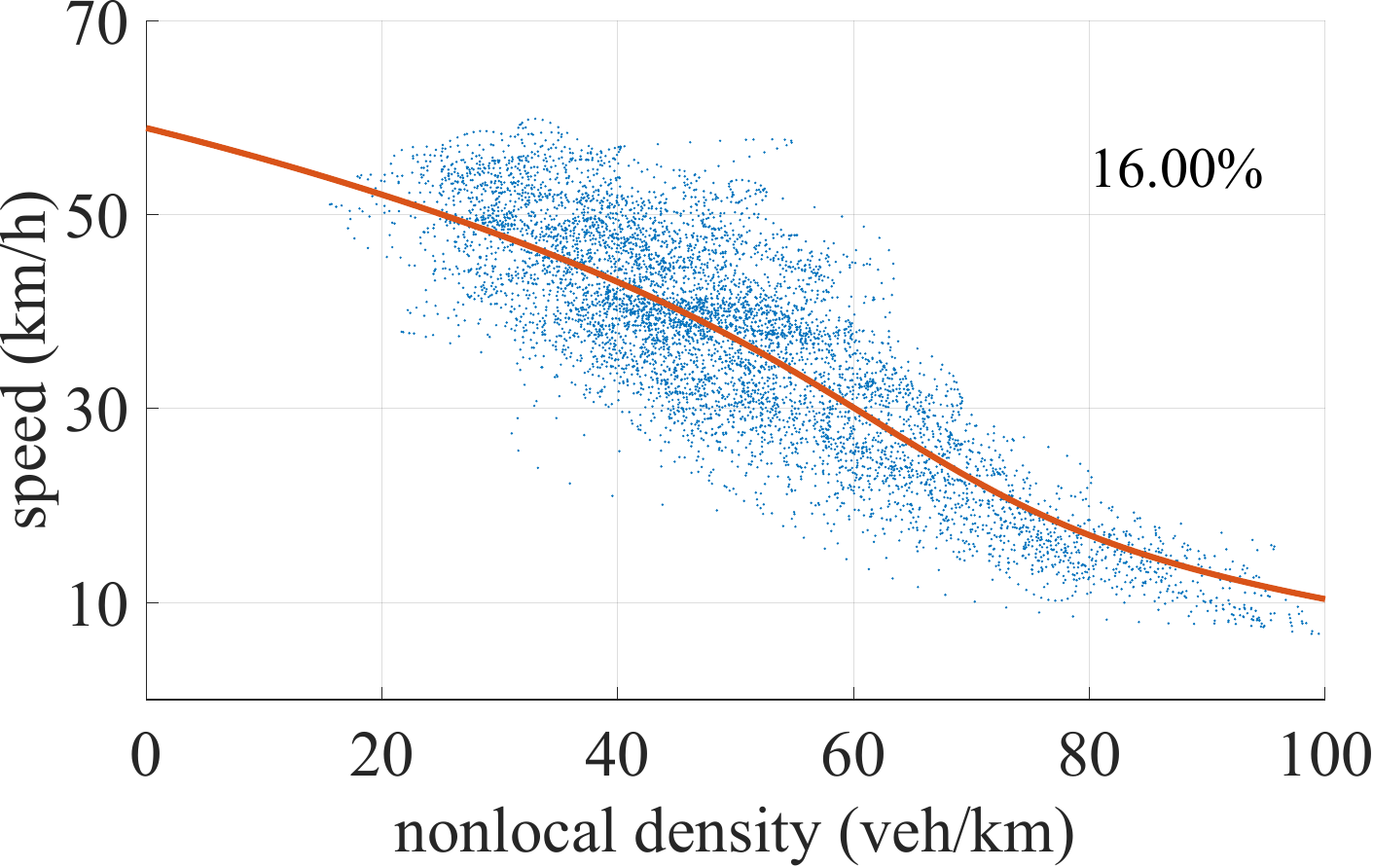}}
    \subfigure[Linear kernel~\eqref{eq:omega decrease}]{\includegraphics[width=0.245\linewidth]{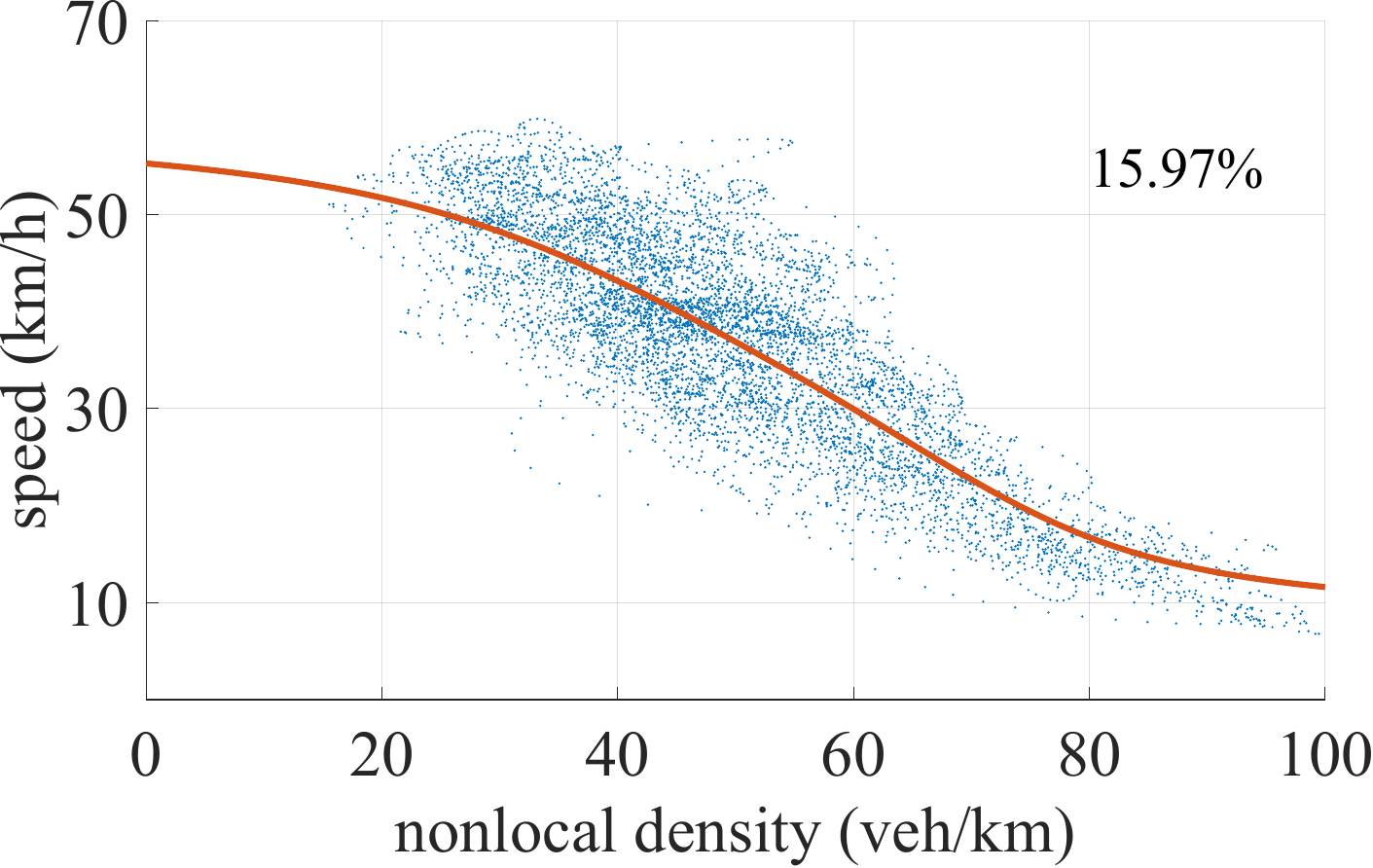}}
    \subfigure[Learned kernel]{\includegraphics[width=0.245\linewidth]{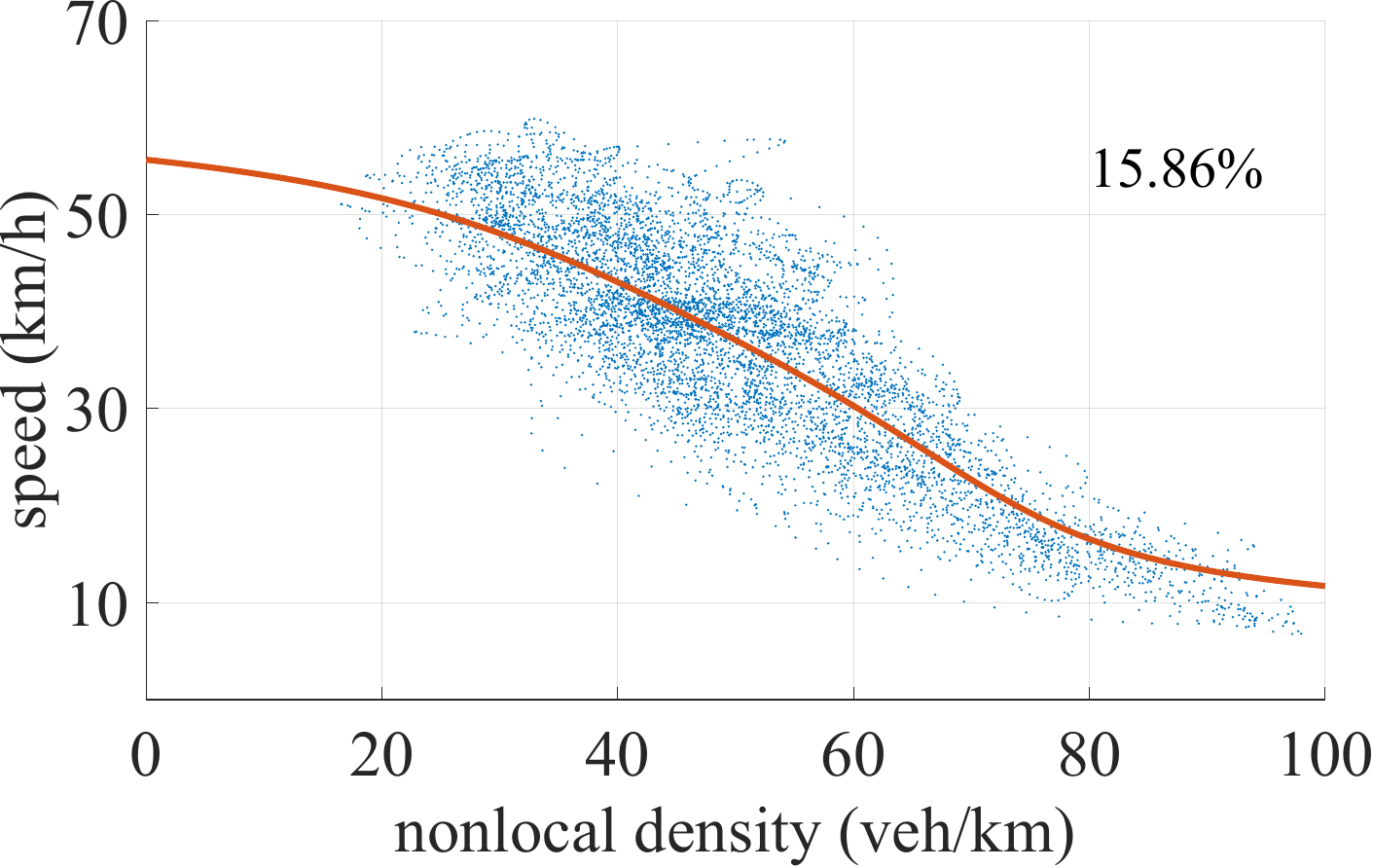}}
    \vspace{-2em}
    \caption{Learned fundamental diagram with local and nonlocal models.}\label{fig:FD}
\end{figure}

\subsection{Nonlocal effect on fundamental diagram}\label{sec:analyze FD}

In Fig.~\ref{fig:FD}, we give the scatter plot of the speed and nonlocal density under different look-ahead kernels with learned fundamental diagram. We also plot the calibrated fundamental diagram. As shown in Fig.~\ref{fig:FD}, in nonlocal LWR,  there still exists a static relationship between speed and nonlocal density. 
To evaluate the calibration accuracy of  fundamental diagram, we use relative RMSE as: 
\begin{equation}
    \setlength{\abovedisplayskip}{\abovegap}
    \setlength{\belowdisplayskip}{\belowgap}
    \mathrm{E}_{v} = \frac{\sqrt{\sum_{i,j} \left( \hat V_{\eta}(\hat \rho_{\eta} (t_i,x_j)) - v(t_i,x_j) \right)^2  }}{\sqrt{\sum_{i,j} \left( v(t_i,x_j) \right)^2}} \times 100\%.
\end{equation}
We give the calibration error in each subfig in Fig.~\ref{fig:FD}. Comparing the fundamental diagram of local model in Fig.~\ref{fig:FD}(a) and nonlocal models in Fig.~\ref{fig:FD}(b)-(d), we see that the nonlocal term only has a marginal effect on the fundamental diagram. For other fundamental diagrams, Greenshields~\eqref{eq:FD greenshields}, UnderWood~\eqref{eq:FD exp}, and Drake~\eqref{eq:FD exp2}, we also find similar results, i.e., the fundamental diagram of nonlocal and local density is approximately the same.

\section{Conclusion}

In this paper, we learn the look-ahead kernel and fundamental diagram of nonlocal LWR model that best fit  ring-road experimental traffic data using physics-informed deep learning. We found out that the traffic data indeed demonstrates look-ahead nonlocal phenomena, and the learned nonlocal LWR model yields a more accurate estimation of traffic wave propagation. Our result also underscored the importance of choosing physics prior in PIDL. For the optimal look-ahead kernel, we found that one to two leader vehicles account for a majority of the anticipation effect in the ring road setting.  For the fundamental diagram, we found that look-ahead has only a marginal effect on the calibrated fundamental diagram. 
The future extension of this paper includes analyzing nonlocal effects in straight roads and control design for nonlocal LWR models.

\clearpage
\bibliography{ref}
\end{document}